\documentclass{article} 
\usepackage[preprint]{colm2026_conference}

\usepackage{microtype}
\usepackage{hyperref}
\usepackage{url}
\usepackage{booktabs}
\usepackage{multirow}
\usepackage{makecell}
\usepackage{graphicx}
\usepackage{caption}
\usepackage{subcaption}

\usepackage{amsmath,amsfonts,bm}

\def\eqref#1{equation~\ref{#1}}

\def\1{\bm{1}}

\DeclareMathAlphabet{\mathsfit}{\encodingdefault}{\sfdefault}{m}{sl}
\SetMathAlphabet{\mathsfit}{bold}{\encodingdefault}{\sfdefault}{bx}{n}

\def\gL{{\mathcal{L}}}

\newcommand{\E}{\mathbb{E}}

\usepackage{lineno}
\usepackage[T1]{fontenc}

\usepackage{fontawesome5}
\usepackage{wrapfig}
\usepackage{tcolorbox}
\usepackage{enumitem}
\usepackage[capitalise]{cleveref}

\definecolor{darkblue}{rgb}{0, 0, 0.5}
\hypersetup{colorlinks=true, citecolor=darkblue, linkcolor=darkblue, urlcolor=darkblue}
\newcommand{\recipe}[0]{Odysseus}

\title{\recipe{}: Scaling VLMs to 100+ Turn Decision-Making in Games via Reinforcement Learning}

\author{Chengshuai Shi$^{1,*}$, Wenzhe Li$^{1,*}$, Xinran Liang$^{1,*}$, Yizhou Lu$^{2}$, Wenjia Yang$^{3}$, \\ \textbf{Ruirong Feng$^{1}$, Seth Karten$^{1}$, Ziran Yang$^{1}$, Zihan Ding$^{1}$, Gabriel Sarch$^{1}$,} \\ \textbf{Danqi Chen$^{1}$, Karthik Narasimhan$^{1}$, Chi Jin$^{1}$}\\
$^1$ Princeton Language and Intelligence, Princeton University \\
$^2$ Fudan University \qquad $^3$ Tsinghua University\\
$^*$ Equal contribution in random order
}

\begin{document}

\ifcolmsubmission
\linenumbers
\fi

\maketitle

\vspace{-0.3in}
\begin{center}
    Project Page: \href{https://odysseus-project.github.io/}{\raisebox{-0.1em}{\faExternalLink*} odysseus-project.github.io/}
\end{center}

\begin{abstract}
Given the rapidly growing capabilities of vision-language models (VLMs), extending them to interactive decision-making tasks such as video games has emerged as a promising frontier. However, existing approaches either rely on large-scale supervised fine-tuning (SFT) on human trajectories or apply reinforcement learning (RL) only in relatively short-horizon settings (typically around 20--30 turns). In this work, we study RL-based training of VLMs for long-horizon decision-making in \textit{Super Mario Land}, a visually grounded environment requiring 100+ turns of interaction with coordinated perception, reasoning, and action. We begin with a systematic investigation of key algorithmic components and propose an adapted variant of PPO with a lightweight turn-level critic, which substantially improves training stability and sample efficiency over critic-free methods such as GRPO and Reinforce++. We further show that pretrained VLMs provide strong action priors, significantly improving sample efficiency during RL training and reducing the need for manual design choices such as action engineering, compared to classical deep RL trained from scratch. Building on these insights, we introduce \recipe{}, an open training framework for VLM agents, achieving substantial gains across multiple levels of the game and at least $3\times$ average game progresses than frontier models. Moreover, the trained models exhibit consistent improvements under both in-game and cross-game generalization settings, while maintaining general-domain capabilities. Overall, our results identify key ingredients for making RL stable and effective in long-horizon, multi-modal settings, and provide practical guidance for developing VLMs as embodied agents.
\end{abstract}

\begin{figure}[hb]
    \centering
    \includegraphics[width=0.75\linewidth]{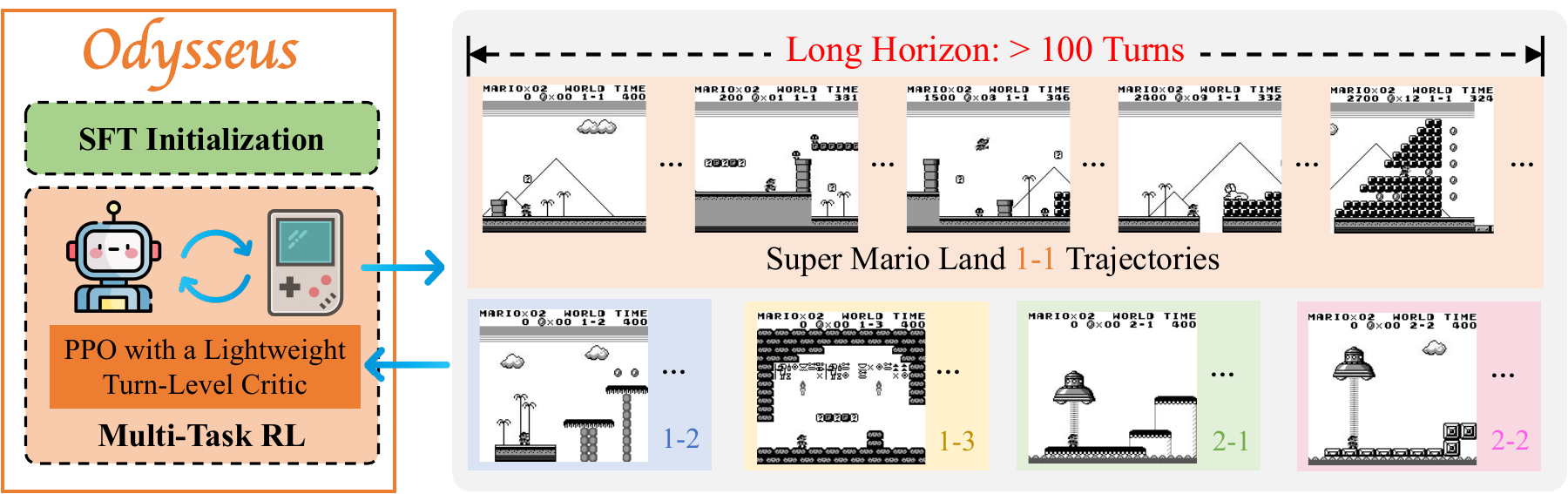}
    \includegraphics[width=0.24\linewidth]{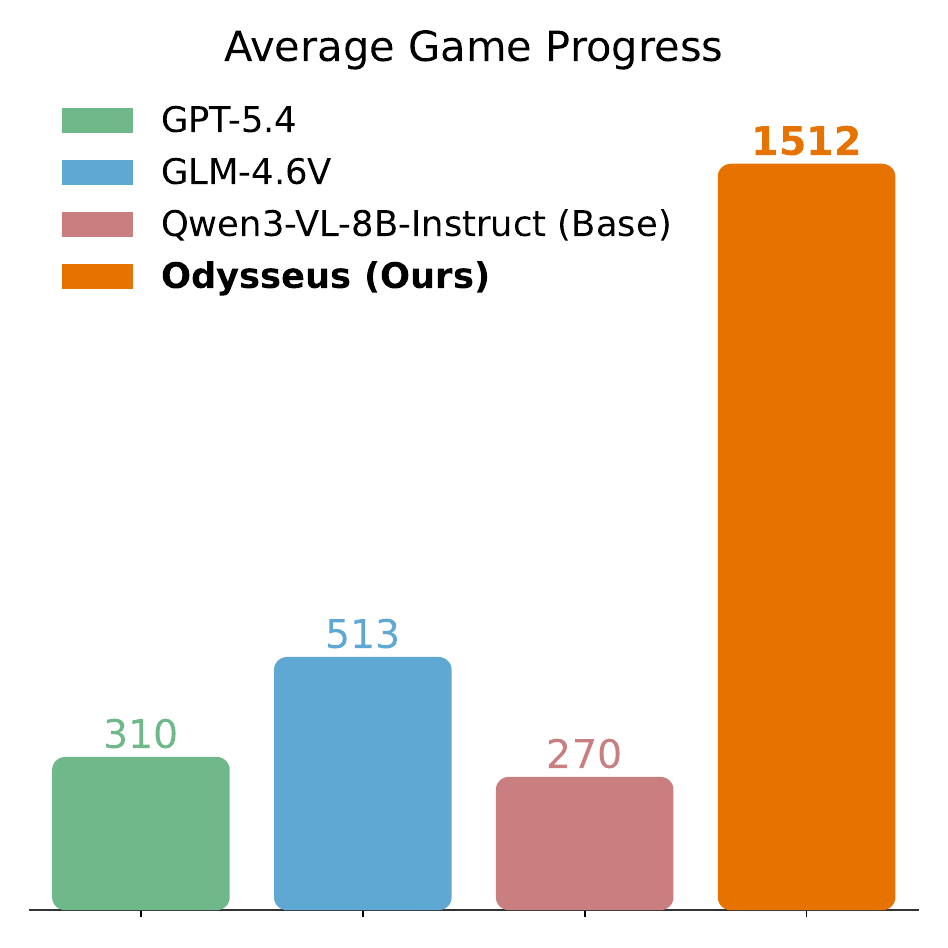}
    \caption{An overview of \recipe{} for scaling VLMs to 100+ turn decision-making in the video game \textit{Super Mario Land}, along with a comparison of performance averaged over the first five game levels across frontier models (GPT-5.4 and GLM-4.6V), the base model (Qwen3-VL-Instruct-8B) prior to training, and the \recipe{} model after training. We observe that \recipe{} achieves approximately $5\times$ higher game progress than GPT-5.4, $3\times$ higher than GLM-4.6V, and $6\times$ higher than the base model.}
    \label{fig:overview}
\end{figure}

\section{Introduction}\label{sec:intro}
\vspace{-0.1in}
Multi-modal foundation models, particularly vision-language models (VLMs), have demonstrated remarkable capabilities across a wide range of domains, such as image captioning, object detection, and visual reasoning. Building on these advances, there is growing interest in extending them toward \emph{agentic tasks}, where models are endowed with the ability to interact with external environments \citep{yao2022react}. Representative applications span web agent \citep{deng2023mind2web}, GUI agents \citep{nguyen2025gui}, and software engineering agent \citep{chan2024mle,yang2024swe}, with increasing attention on \textbf{embodied agents} performing interactive decision-making tasks in physically grounded or simulated environments \citep{driess2023palm,zitkovich2023rt,mu2023embodiedgpt,raad2024scaling,szot2025multimodal}. 

As a long-standing testbed of simulated embodied task \citep{mnih2015human,vinyals2019grandmaster}, video games have attracted growing interest for evaluating, scaffolding, and training VLM agents \citep{zhang2025videogamebench,hu2025lmgame,wang2023voyager,karten2025pokeagent,bolton2025sima}. From the perspective of training, while reinforcement learning (RL) has achieved great success in training classical deep neural networks in video games \citep{badia2020agent57}, and has recently been applied to improve foundation models in domains such as preference alignment \citep{ouyang2022training} and reasoning \citep{guo2025deepseek}, existing approaches to fine-tuning VLMs for embodied tasks—particularly in video games—remain limited. Current methods either rely on large-scale supervised fine-tuning (SFT) with human trajectories (i.e., imitation learning) \citep{tan2025lumine,magne2026nitrogen}, which is difficult to scale, or apply RL only to relatively short-horizon tasks (typically around 20--30 turns) \citep{zhai2024fine,wang2025vagen}. It remains unclear whether RL can be effectively applied for training VLMs in more challenging, long-horizon ($> 100$ turns) decision-making tasks.

In this work, we use the video game \textit{Super Mario Land} to study this regime. Despite its simplicity, this environment remains challenging even for frontier models \citep{zhang2025videogamebench}. Successful performance requires coordinated perception, reasoning, and action over extended trajectories, often exceeding \textbf{100 turns}, and the ability to generalize across diverse levels with varying layouts and dynamics. Our contributions are summarized as follows.

$\bullet$ \textbf{Algorithmic ingredients.} We investigate key algorithmic choices required to effectively fine-tune VLMs via RL in long-horizon game environments. While popular critic-free methods perform poorly in this setting, we demonstrate that an adapted version of PPO achieves strong stability and sample efficiency. Crucially, we introduce a \emph{lightweight turn-level critic} (instead of a large model based one as in \citet{wang2025vagen,zhai2024fine}) and \emph{positive-advantage filtering}, which together decouple temporal credit assignment from token generation, mitigate optimization instability, and bypass the massive computational overhead usually associated with large-model-based actor-critic training.

$\bullet$ \textbf{VLM-based RL training v.s. classical deep RL.} Beyond enabling stable RL training of VLMs in this setting, we identify the advantages of VLM-based RL compared to classical deep RL which trains policies from scratch. We show that pretrained VLMs offer strong action priors that improve sample efficiency and reduce the need for manual designs such as action space engineering. This highlights the importance of general-purpose knowledge encoded in foundation models for scaling toward capable embodied agents.

$\bullet$ \textbf{An open training framework for practical agentic tasks.} Built upon these insights, we introduce \textbf{\recipe{}}, an open and practical training framework (\cref{fig:overview}) that integrates lightweight supervised initialization with multi-task RL. We show that \recipe{} enables stable training over tens of millions of interaction samples, achieving substantial performance gains across the game over the base model, and outperforming both open-source and proprietary frontier models by a large margin (at least $3\times$ improvement in game progress). Furthermore, the resulting agents exhibit generalization both within the game and to related game environments, while retaining their capabilities on general-domain multi-modal tasks.

Taken together, our results demonstrate that RL can be made stable and effective for training VLMs in \textbf{100+ turn decision-making environments}. Moreover, once RL is properly stabilized, foundation models provide strong priors that further facilitate learning. We hope this work provides a practical foundation and opens the door to future advances in RL training of multi-modal foundation models as embodied agents.

\section{Related Work}
\label{sec:related_work}

\textbf{Games and Simulated Environments.} 
While video games and simulated environments have long served as testbeds for machine learning, the recent exploration of RL for VLMs is mostly focused on short-horizon scenarios, such as AlfWorld~\citep{ALFWorld20}, Sokoban, and FrozenLake~\citep{wang2025vagen}. In contrast, we focus on the video game \textit{Super Mario Land} as a compact but appealing testbed for VLMs in long-horizon embodied control. It imposes substantially richer spatial grounding and closed-loop control than short-horizon gridworld-style tasks, while remaining lightweight and easy to scale for controlled studies compared with large open-world simulators~\citep{fan2022minedojo,tan2025lumine}.

\textbf{Foundation Models for Decision-Making.}
Recent advancements have shifted toward fine-tuning pretrained foundation models directly for embodied control, yielding capable agents in robotic manipulation~\citep{black2024pi_0,liu2024rdt} and cross-game generalization~\citep{tan2025lumine,bolton2025sima}. However, these approaches heavily depend on Supervised Fine-Tuning (SFT) with large amounts of action-labeled demonstration data. Our work differentiates itself by focusing on the RL perspective, investigating how to effectively adapt foundation models without relying on extensive SFT data.

\textbf{RL for Foundation-Model Agents.}
A growing body of work studies RL for multi-turn language and vision-language agents~\citep{zhai2024fine,wang2025vagen,wang2025ragen,li2026salt,he2026hierarchy}. These methods often introduce specialized machinery for trajectory decomposition, token-level advantage estimation, or hierarchical credit assignment, and are typically evaluated on environments with relatively short horizons (20--30 turns). In contrast, we focus specifically on long-horizon, visually grounded embodied environments that require 100+ turns of interaction with chain-of-thought (CoT) reasoning. Through rigorous ablations, we show that a comparatively simple PPO-based approach with the right critic design is sufficient to make RL stable and effective.

We refer readers to \cref{app:extended_related_work} for a more comprehensive literature review.

\section{VLMs for Decision-Making in \textit{Super Mario Land}}
\subsection{Formulation}
A commonly adopted abstraction for decision-making tasks follow the formulation of Partially Observable Markov Decision Process (POMDP) from classical RL literature~\citep{sutton1998reinforcement}. Specifically, a POMDP is defined by the tuple $\langle \mathcal{S}, \mathcal{A}, \Omega, \mathcal{P}, \mathcal{O}, \mathcal{R}\rangle$, where $\mathcal{S}$ denotes the underlying state space, $\mathcal{A}$ is the action space, and $\Omega$ represents the observation space. At each turn $t$, the environment is in an unobserved state $s_t \in \mathcal{S}$, while the agent receives an observation $o_t \in \Omega$. The agent takes an action $a_t \in \mathcal{A}$, after which the environment transitions to a new state $s_{t+1} \sim \mathcal{P}(\cdot \mid s_t, a_t)$, and produces a new observation $o_{t+1} \sim \mathcal{O}(\cdot \mid s_{t+1})$ along with a scalar reward $r_t = \mathcal{R}(s_t, a_t)$.

The agent aims to achieve high performance in the environment, which in RL is formulated as maximizing the expected cumulative discounted reward under a parameterized policy $\pi_\theta$ (i.e., the adopted VLM), i.e., $\mathbb{E}_{\pi_\theta}\left[\sum_{t=0}^{T-1} \gamma^t r_t\right]$, with $T$ denoting the horizon and $\gamma \in [0, 1)$ denoting the discount factor. In the partially observable setting, the policy is defined as a mapping from the interaction history $h_t = (o_1, a_1, \dots, o_t)$ to a distribution over actions, i.e., $a_t \sim \pi_\theta(\cdot \mid h_t)$, where the exact dependence on $h_t$ depends on the design of the agent.

\subsection{An Overview of \textit{Super Mario Land}} 

\begin{table}[thb]
    \centering
    \begin{tabular}{cc}
\hline
Game & Effective Horizon \\
\hline
AlfWorld &  $\sim 10-20$\\
Sokoban ($6\times 6$) &  $\sim 5-30$ \\
Frozen Lake ($4\times 4$) &  $\sim 5-30$\\
Super Mario Land & $> 100$\\
\hline
\end{tabular}
\caption{Effective horizon comparisons.}
    \label{tbl:horizon}
    \vspace{-0.2in}
\end{table}

To study the problem of effectively training VLMs for long-horizon decision-making tasks via RL, we consider the video game \textit{Super Mario Land} as a compelling testbed. The game requires agents to perform accurate spatial perception and reasoning over extended trajectories (often $100+$ turns), together with precise motor control to navigate diverse levels populated with obstacles and adversaries. \cref{tbl:horizon} highlights the significantly longer horizon of \textit{Super Mario Land} compared with previous game environments used in VLM-RL literature \citep{ALFWorld20,wang2025vagen}. This game consists of 12 levels in total (in particular, $4$ worlds, each with $3$ levels), and we mainly consider $10$ of these levels in this work, excluding $2$ levels (World 2 Level 3 and World 4 Level 3) due to their distinct control mechanisms.

Recent works~\citep{park2025orak, zhang2025videogamebench, hu2025lmgame} have also used this game or other versions from the Super Mario series to benchmark foundation models. As demonstrated in these works—and as we will show later—even state-of-the-art models struggle in the zero-shot setting. For example, tasks such as jumping over a gap or avoiding a moving threat at the correct timing remain challenging, resulting in brittle policies that rarely progress beyond the initial stages of the game. In contrast, a human player with no prior experience can readily achieve non-trivial progress.

With the interaction protocols specified in \cref{sec:protocol}, we briefly note that for the game, the \textbf{state space} $\mathcal{S}$ corresponds to the full internal state of game RAM. As in human gameplay, the agent does not have access to this state, and instead observes the \textbf{observation space} $\Omega$, consisting of rendered pixel frames and textual prompts. The \textbf{action space} $\mathcal{A}$ is a discrete set of combinations derived from standard controller inputs. For language-based foundation models, actions are produced by generating text tokens that specify the buttons to be pressed. Finally, the \textbf{reward function} $\mathcal{R}$ reflects task progress, and is primarily defined in this work as forward movement at each turn, i.e., $r_t = x_{t+1} - x_t$, a minimal yet dense learning signal for task progress,
where $x_t$ is the $x$-coordinate of Mario at turn $t$ read from game RAM.

\subsection{Interaction Protocol}
\label{sec:protocol}

\begin{figure}[tb]
    \centering
    \includegraphics[width=0.8\linewidth]{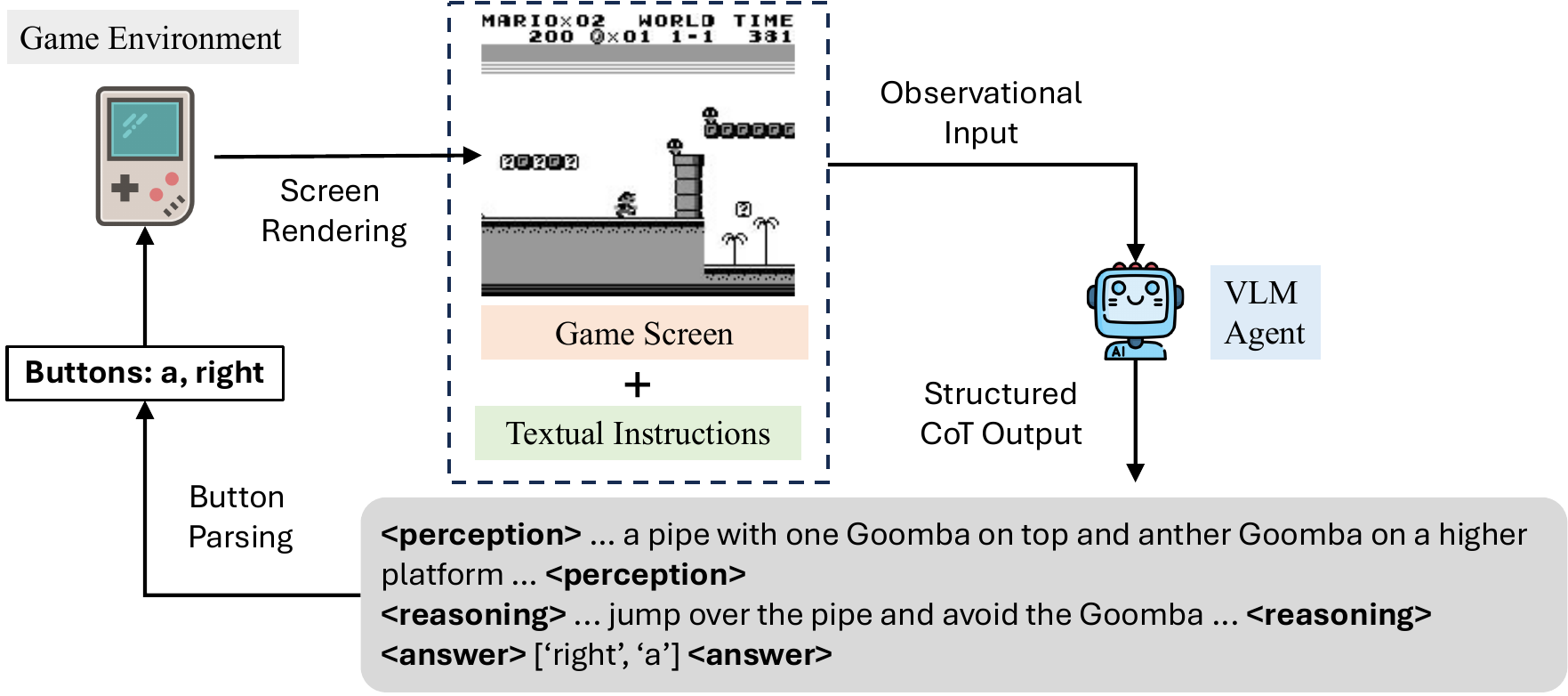}
    \caption{The interaction protocol between the VLM agent and the game environment.}
    \vspace{-0.2in}
    \label{fig:interaction_protocol}
\end{figure}

To facilitate the interaction between an VLM-based agent and the dynamic environment of \textit{Super Mario Land}, we establish a structured, turn-based protocol, as illustrated in \cref{fig:interaction_protocol}:

\textbf{Observational Inputs.}
The agent receives a comprehensive prompt specifying the overall rules of the environment. This includes the basic mechanics and objectives of the game, a precise definition of the available discrete action space, and output format instructions. The full text of this prompt is provided in \cref{app:interaction}. In addition to these general instructions, the agent is given the current game frame rendered on the screen. While recent agent designs often provide richer inputs to the VLM (e.g., a longer observation history, or information parsed from underlying game states)~\citep{park2025orak}, we intentionally adopt a minimal-scaffolding design by providing only the current game frame and the prompt.

\textbf{Structured Chain-of-Thought (CoT).}
To elicit robust spatial-temporal grounding, we instruct the VLM to structure its decision process using three XML-style tags:
\begin{itemize}[noitemsep, topsep=0pt, leftmargin=*]
    \item \textbf{\texttt{<perception>}:} The agent first explicitly describes the visual state of the screen. This grounding step encourages the model to identify Mario's location, nearby obstacles, enemies, and interactive elements such as coins or pipes.
    \item \textbf{\texttt{<reasoning>}:} The agent then lays out its strategy step by step. It explains the actions required to respond to the current state, such as timing a jump to collect floating coins or moving right to approach a pipe stack.
    \item \textbf{\texttt{<answer>}:} Finally, the agent outputs the selected action combination as a list of button strings (e.g., \texttt{[`a', `right']}). The action space permits pressing up to two buttons simultaneously out from a total list of seven (i.e., \texttt{a}, \texttt{b}, \texttt{up}, \texttt{down}, \texttt{left}, \texttt{right}, \texttt{noop}), enabling more complex behaviors such as running jumps.
\end{itemize}

This response format largely follows the style of ReAct \citep{yao2022react} and is also similarly adopted in \citet{chen2025g1,wang2025vagen}.

\textbf{Action Execution.}
Once the structured CoT is generated by the VLM, the final action is parsed and executed in the game environment. Because a single emulator frame ($1/60$ s) produces negligible movement, we implement a frame-skip mechanism: the chosen discrete action is repeatedly applied for a fixed number of consecutive frames to ensure an observable effect. Details are provided in \cref{app:interaction}.

\section{Algorithmic Ingredients of Stable RL for VLMs}\label{sec:rl_comparison}
In this section, we focus on examining whether RL can be effectively applied to train VLMs in the considered long-horizon game and if so, what algorithmic designs are important. 

\subsection{PPO with a Lightweight Turn-Level Critic}
\begin{figure}[tb]
    \centering
    \includegraphics[width=\linewidth]{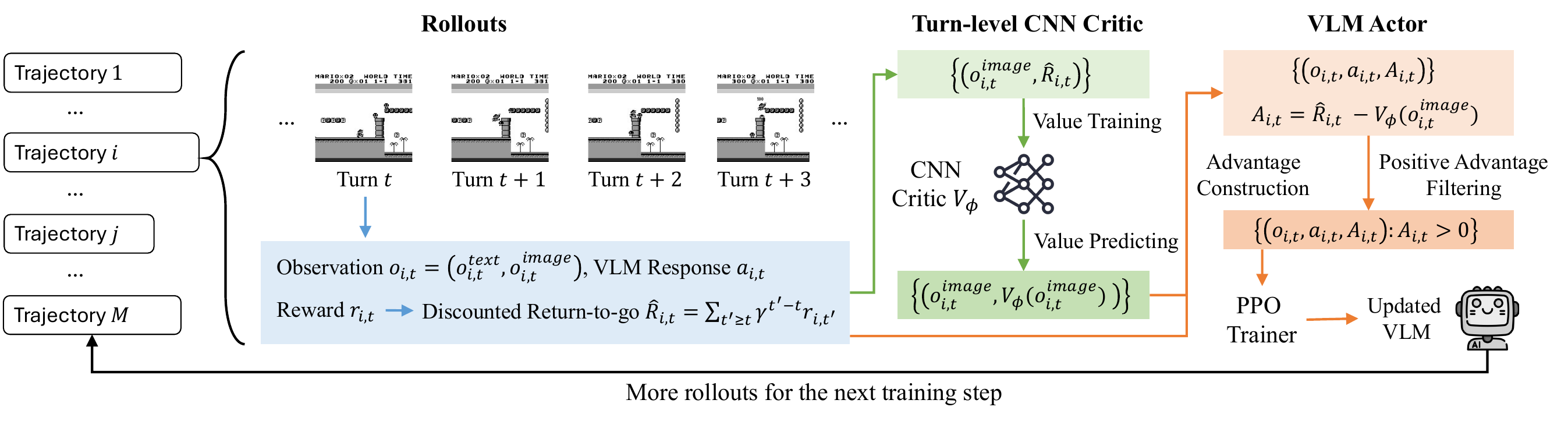}
    \caption{The adapted PPO algorithm used in \recipe{} with a lightweight turn-level CNN critic and positive advantage filtering.}
    \vspace{-0.2in}
    \label{fig:ppo}
\end{figure}

We first consider extending commonly-used RL fine-tuning algorithms to the considered long-horizon dense-reward settings, including GRPO \citep{shao2024deepseekmath} and Reinforce++ \citep{hu2025reinforce++}, which leverage critic-free strategies for advantage estimation. However, both outcome-reward and process-reward variants of these methods fail to learn effective policies that can make consistent multi-step progress (see results in \cref{subsec:rl_alg_comp}). This particular failure mode motivates us to revisit the classical Proximal Policy Optimization (PPO) algorithm~\citep{schulman2017proximal}, which employs a learned critic that enables better long-term credit assignment and low-variance advantage estimation~\citep{schulman2015high}. We defer full algorithm details to \cref{app:vlm_rl} while highlighting several key design choices as follows, which are also illustrated in \cref{fig:ppo}.

One major challenge in using PPO to train foundation models is the computation overhead of learning the critic --- prior work usually learns a large-model-based token-level critic, which nearly doubles the memory and computation costs compared to critic-free methods. To address this issue, we propose two key changes to the original PPO algorithm used in RLHF~\citep{ouyang2022training}. 
First, a \emph{turn-level critic} is adopted. Compared to token-level critic learning, turn-level critic can directly leverage the dense reward signals from game environments and address the long-horizon temporal credit assignment more effectively.

Furthermore, we demonstrate that the critic network can be designed as a remarkably lightweight module, particularly for environments characterized by rich visual state information. Rather than employing a computationally expensive, secondary VLM as the value network, we show that Convolutional Neural Network (CNN) critics, which are sufficient for classical deep RL~\citep{schulman2017proximal,raffin2021stable,huang202237}, can already effectively stabilize training and lead to improved performance. These two designs together yield an important insight with broad implications for scaling VLM RL to long-horizon tasks: by delegating turn-level value estimation to a small module, we can drastically reduce the memory and computational bottlenecks associated with large-scale actor-critic training, making RL fine-tuning significantly more efficient and accessible.

Moreover, we consider an additional algorithmic modification that filters out samples with negative advantages (i.e., $\hat{A}_t < 0$) during training, effectively clipping the advantage at zero, which is referred to as ``positive-advantage filtering'' in subsequent discussions. This design is motivated by empirical observations that negative-advantage samples can destabilize optimization, a phenomenon reported in both foundation model fine-tuning \citep{xiong2025minimalist,deng2025effect} and classical deep RL \citep{hamalainen2020ppo,jesson2023relu}. We note that the role of positive versus negative advantage samples in RL training remains an actively studied question \citep{deng2025effect,carrino2026complicated,zhu2025surprising}. Our results provide additional empirical evidence that may help shed light on this direction.

\subsection{Comparisons with GRPO and Reinforce++} 

\label{subsec:rl_alg_comp}
To better understand different RL algorithmic components, we conduct a controlled experiment on a challenging scenario (i.e., the scenario shown in \cref{fig:interaction_protocol}) in World 1 Level 1 of \textit{Super Mario Land}. The task is to progress as far as possible without dying, beginning with the immediate challenge of jumping over a tall pipe while avoiding two approaching enemies, followed by additional obstacles including platform gaps and additional enemies.

\begin{figure}[t]
\centering
\hfill
\begin{subfigure}{0.49\linewidth}
\centering
\includegraphics[width=\linewidth]{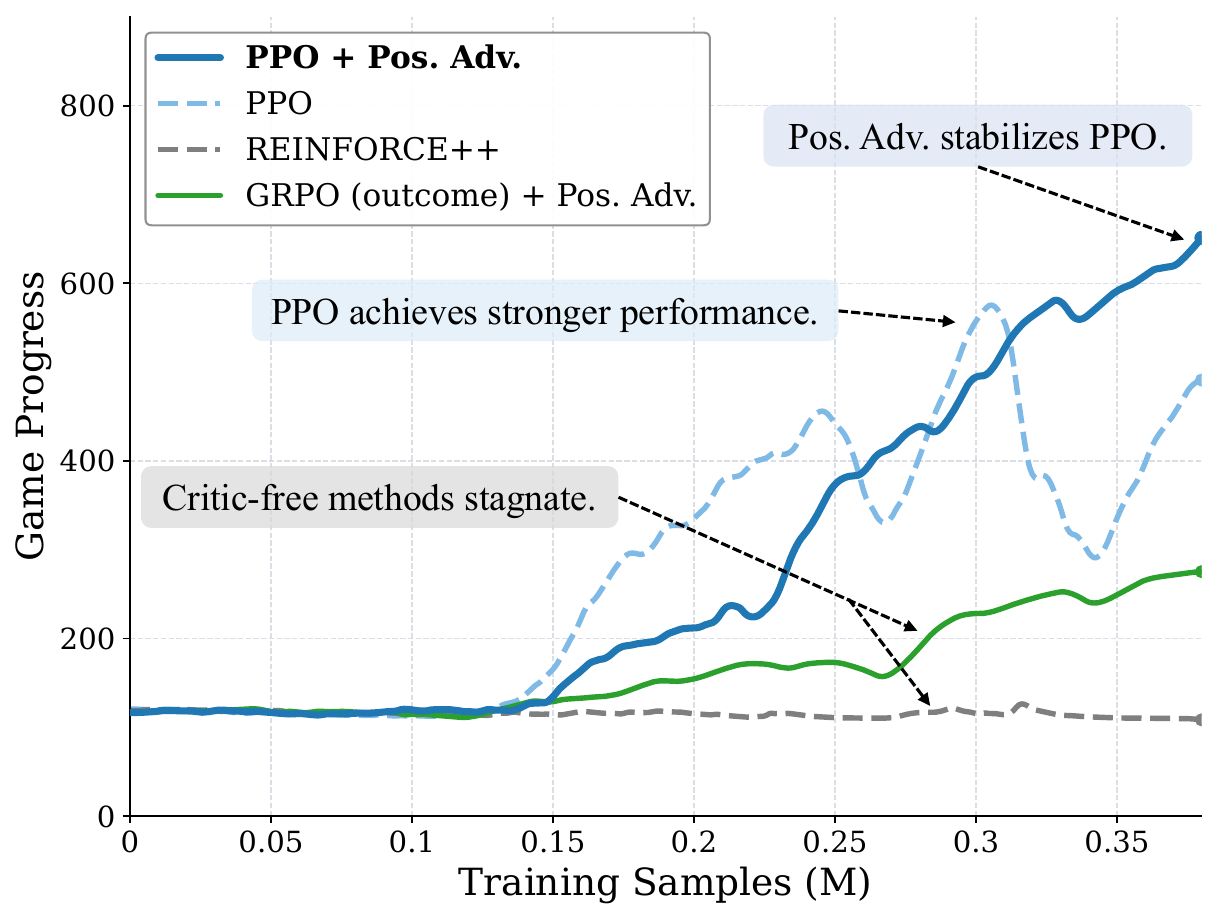}
\caption{Different VLM-based RL methods.}
\label{fig:compare_vlm}
\end{subfigure}
\hfill
\begin{subfigure}{0.49\linewidth}
\centering
\includegraphics[width=\linewidth]{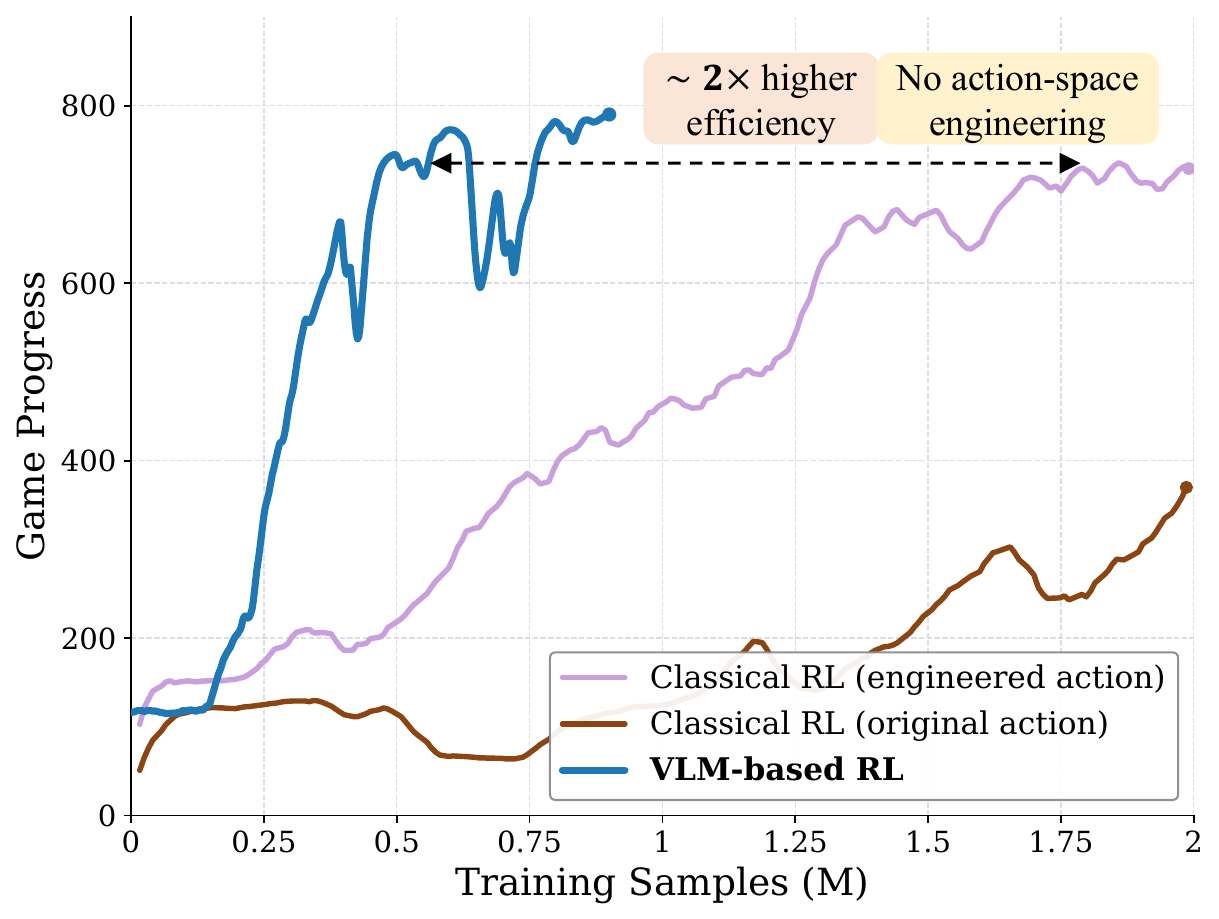}
\caption{VLM-based RL v.s. classical deep RL.}
\label{fig:compare_classical}
\end{subfigure}
\hfill
\caption{(a) Comparison of VLM-based RL training methods with training samples limited to $0.38$M. PPO with a turn-level CNN critic substantially outperforms critic-free methods, and positive-advantage filtering further stabilizes training. (b) Comparison between VLM-based RL (PPO with a turn-level CNN critic and positive-advantage filtering) and classical RL (PPO training a CNN policy from scratch). VLM-based RL achieves roughly $2\times$ higher sample efficiency, even without action-space engineering. Curves are averaged over at least two independent runs and plotted via EMA smoothing with a factor $0.85$; individual runs and additional methods are provided in \cref{fig:vlm_compare_full} and \cref{fig:classical_rl_compare_full}.}
\vspace{-0.2in}
\end{figure}

We compare the following candidate methods: (i) GRPO with outcome rewards; (ii) GRPO with outcome rewards and positive-advantage filtering; (iii) GRPO with process rewards; (iv) GRPO with process rewards and positive-advantage filtering; (v) Turn-level Reinforce++; (vi) PPO with a learned CNN critic; and (vii) PPO with a learned turn-level CNN critic and positive-advantage filtering. Their detailed implementations are provided in \cref{app:vlm_rl}. We report performance as a function of the number of training samples in \cref{fig:compare_vlm} (with the full results in \cref{fig:vlm_compare_full}), where Qwen3-VL-8B-Instruct \citep{bai2025qwen3} is the base model for training.

Overall, critic-free methods (GRPO and Reinforce++) exhibit unstable learning dynamics and limited performance gains, regardless of reward design or the use of advantage filtering. Particularly, only GRPO with outcome rewards leads to observable improvements after training. In contrast, PPO-based methods achieve substantially stronger and more stable improvements, underscoring the importance of a learned critic for effective credit assignment in this long-horizon setting. Furthermore, positive-advantage filtering improves training stability when combined with PPO. This result highlights the importance of algorithmic components for making RL stable and effective for training VLMs on long-horizon tasks.

\section{VLM-Based RL Training versus Classical Deep RL}\label{sec:classical_rl}
With stable RL for training VLMs in long-horizon game environments established in \cref{sec:rl_comparison}, a natural follow-up question is whether this approach offers tangible advantages over classical deep RL methods, which are also capable of solving similar tasks. In this section, we study this question through the lens of sample efficiency.

\textbf{Hypothesis: VLM-based RL is more sample-efficient than classical deep RL trained from scratch.} Sample efficiency has long been a central challenge in classical deep RL, especially in complex visual environments~\citep{mnih2015human,badia2020agent57,vinyals2019grandmaster}. A key reason is that such agents must learn both perception and control from scratch, without access to prior knowledge of visual semantics or action dynamics. In contrast, pretrained VLMs already encode rich visual representations and broad world knowledge, providing strong priors for both perception and action. From this perspective, VLM-based RL can be viewed as narrowing the gap between conventional RL agents and human gameplay, where prior knowledge plays a crucial role. We therefore hypothesize that VLM-based RL achieves higher sample efficiency than classical deep RL methods trained from scratch.

\textbf{Experimental results: VLM-based RL achieves higher sample efficiency with less manual designs.} To test this hypothesis, we evaluate classical deep RL on the same task used in \cref{subsec:rl_alg_comp}. We adopt PPO with a CNN policy as the baseline, given its widespread use and strong empirical performance in prior work. We consider two action-space designs: (i) an \emph{original} action space with $22$ actions, covering all valid button combinations of up to two simultaneous presses, and (ii) an \emph{engineered} action space with $8$ button combinations designed to better reflect human gameplay. Implementation details are provided in \cref{app:classical_rl}. Notably, this action-space engineering is applied only to classical deep RL; the VLM agent follows the interaction protocol described in \cref{sec:protocol}.

As shown in \cref{fig:compare_classical}, PPO with the original action space makes only slow progress, likely because exploration is difficult in a large combinatorial action space. Using the engineered action space substantially improves performance by restricting the policy to a smaller and more semantically meaningful set of actions. However, even with this manual design, classical deep RL remains significantly less sample-efficient than VLM-based RL (i.e., PPO with positive-advantage filtering; see \cref{subsec:rl_alg_comp}), requiring roughly $2\times$ more samples to reach a comparable converged maximum performance.

These results support our hypothesis that pretrained VLMs provide strong inductive biases that reduce the exploration burden in long-horizon tasks. More broadly, they suggest that pretrained VLMs can serve as knowledgeable priors for RL, improving sample efficiency while reducing the need for manual engineering in embodied decision-making problems.

\section{\recipe{}: An Open and Practical Training Framework}
With the key algorithmic ingredients identified in \cref{sec:rl_comparison} and the benefits of training a VLM-based agent established in \cref{sec:classical_rl}, we further extend the scope to finetune a VLM on \textbf{multiple levels} of the game simultaneously. We present \recipe{}, an open framework for training practical decision-making agents, which integrates supervised fine-tuning initialization and multi-task RL training into a pipeline. While our primary instantiation is in \textit{Super Mario Land}, \recipe{} is sufficiently general to inform a broader range of settings.

\subsection{Supervised Initialization}

Our initial experiments showed that the currently available small open-source VLMs (e.g., Qwen3-VL-8B-Instruct) sometimes lack sufficient domain knowledge and perceptual grounding in \textit{Super Mario Land}, likely because such environments are underrepresented in their pre-training data (which typically contains limited coverage of games). For example, they may struggle to distinguish Mario from enemies or to accurately identify their spatial positions. We therefore begin with a light supervised fine-tuning (SFT) stage to inject domain-specific knowledge and improve environment-specific perception.

In particular, we first curate a dataset covering diverse scenarios of \textit{Super Mario Land}. Specifically, we sample around 5,000 frames across the 10 considered levels from two walkthrough videos that complete the game. For each frame, a stronger model (in our case, GPT-o3) is used to generate teacher responses following the same format described in \cref{sec:protocol}, including structured \texttt{<perception>}, \texttt{<reasoning>}, and \texttt{<answer>} fields. We qualitatively verify that these CoT annotations are of consistently high quality in terms of both game knowledge and visual perception. Using the sampled images and generated responses as training data, we then perform standard SFT with cross-entropy loss, while preserving the same input-output format as in \cref{sec:protocol}.

It is worth noting that the goal of this SFT stage is intentionally lightweight compared with previous works \citep{tan2025lumine,bolton2025sima}: it focuses on improving domain knowledge and environment perception, rather than optimizing action control, which is deferred to RL. Accordingly, the curated dataset in this work is first significantly smaller in scale. Also, instead of relying on expert trajectories with annotated actions, the sampled frames from walkthrough videos do not inherently provide action labels. These actions are instead generated by the teacher model, which, while strong in perception and reasoning, does not necessarily produce optimal action decisions, as we will demonstrate later. Nevertheless, as shown in our experiments, this lightweight SFT stage improves the effectiveness of subsequent RL training. Furthermore, since game-play videos are far more readily available than collecting expert trajectories with actions, this approach is inherently more scalable.

\subsection{Reinforcement Learning with Multi-Task Auto-Curriculum}

\begin{wrapfigure}{r}{0.4\textwidth}
\vspace{-0.1in}
    \centering
    \includegraphics[width=\linewidth]{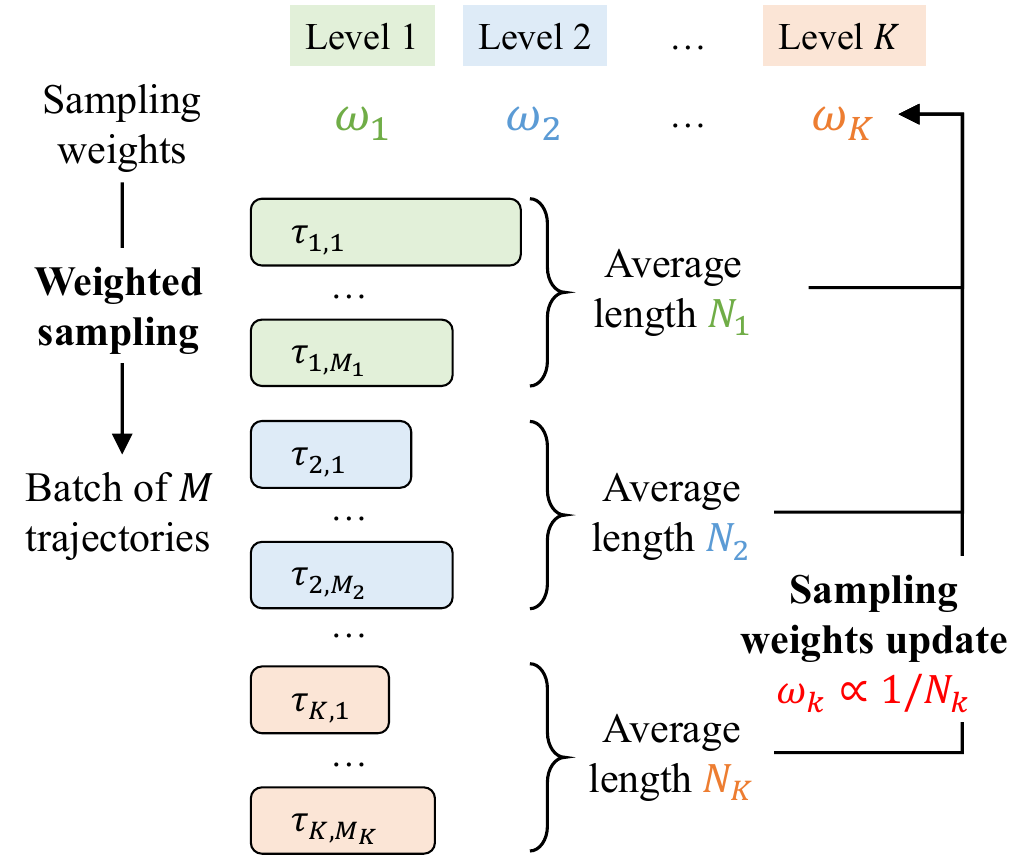}
    \caption{Auto-Curriculum.}
    \label{fig:auto_curriculum}
    \vspace{-0.2in}
\end{wrapfigure}

Building on the environment knowledge and perception capabilities acquired during SFT, we further apply RL to optimize action selection in the environment, thereby improving final performance. Based on the algorithmic findings in \cref{sec:rl_comparison}, we adopt the adapted PPO together with positive-advantage filtering.

To enable multi-task training, each training batch contains trajectories collected from multiple levels of the game. We further introduce an \textbf{auto-curriculum} mechanism to balance learning progress across tasks. Suppose there are $K$ levels involved in training. A batch containing $M$ trajectories in total is denoted by $\mathcal{D} = \bigcup\nolimits_{k \in [K]} \{\tau_{k,m} : m \in [M_k]\}$,
where $\tau_{k,m}$ denotes the $m$-th trajectory from level $k$, and $M_k$ is the number of trajectories sampled from level $k$. A naive strategy is to sample levels uniformly in each batch. However, because different levels can vary substantially in difficulty, uniform level sampling can lead to an undesirable imbalance in the optimization objective. In particular, trajectories from easier levels tend to be longer, since the agent can naturally survive and progress further. For example, if level 1 is easier than level 2, then trajectories in $\{\tau_{1,m} : m \in [M_1]\}$ will typically be much longer than those in $\{\tau_{2,m} : m \in [M_2]\}$. Under uniform level sampling, i.e., $M_1 \approx M_2$, the PPO objective—which aggregates losses over all samples—will therefore contain many more samples from level 1 than from level 2. As a result, optimization becomes biased toward easier levels, potentially at the expense of performance on harder ones.

To promote balanced learning across levels, we introduce an auto-curriculum mechanism based on inverse trajectory weighting~\citep{li2024fightladder}, as demonstrated in \cref{fig:auto_curriculum}. For each level $k$ present in the current batch, let $N_k$ denote the average trajectory length from that level: $N_k = \frac{1}{M_k}\sum\nolimits_{m \in [M_k]} \mathrm{len}(\tau_{k,m})$.
Then, for the next batch, levels are sampled according to $w_k \propto 1/N_k, \forall k \in [K]$.
Intuitively, this inverse weighting up-weights levels with shorter trajectories and down-weights those with longer ones. As a result, it approximately balances the number of training samples contributed by each level. This auto-curriculum therefore shifts training toward less-explored levels in a dynamic manner, improving both sample efficiency and training stability.

\begin{table}[tb]
\setlength{\abovecaptionskip}{0.2em}
\begin{center}
\resizebox{\textwidth}{!}{
\begin{tabular}{c|cccccc}
\toprule
\multirow{2}{*}{Models} &\multicolumn{6}{c}{\bf Average Progress by (World, Level)}\\
\cmidrule{2-7}
& (1,1) & (1,2) & (1,3) & (2,1) & (2,2) & Avg. \\
\midrule
GPT-5.4 & 403.62 {\tiny $\pm$ 78.70} & 67.62 {\tiny $\pm$ 19.84} & 654.86 {\tiny $\pm$ 96.57} & 252.75 {\tiny $\pm$ 27.31} & 173.50 {\tiny $\pm$ 13.15} & 310.47 {\tiny $\pm$ 25.95} \\
Gemini-3-Flash & 529.12 {\tiny $\pm$ 21.51} & 255.88 {\tiny $\pm$ 97.11} & 493.43 {\tiny $\pm$ 48.49} & 187.75 {\tiny $\pm$ 24.90} & 239.50 {\tiny $\pm$ 21.57} & 341.14 {\tiny $\pm$ 23.09} \\
Claude-Sonnet-4.6 & 608.12 {\tiny $\pm$ 66.45} & 132.00 {\tiny $\pm$ 58.28} & 502.62 {\tiny $\pm$ 58.61} & 206.50 {\tiny $\pm$ 18.64} & 291.75 {\tiny $\pm$ 34.40} & 348.19 {\tiny $\pm$ 22.61} \\
\midrule
Qwen3-VL-235B-A22B-Instruct & 424.42 {\tiny $\pm$ 13.26} & 51.69 {\tiny $\pm$ 2.56} & 511.48 {\tiny $\pm$ 15.58} & 186.88 {\tiny $\pm$ 5.40} & 222.77 {\tiny $\pm$ 4.50} & 279.45 {\tiny $\pm$ 4.36} \\
InternVL3.5-241B-A28B & 442.41 {\tiny $\pm$ 14.01} & 78.12 {\tiny $\pm$ 6.57} & 390.11 {\tiny $\pm$ 13.87} & 188.61 {\tiny $\pm$ 5.04} & 196.40 {\tiny $\pm$ 4.32} & 259.13 {\tiny $\pm$ 4.36} \\
GLM-4.6V (106B-A12B) & 731.28 {\tiny $\pm$ 15.15} & 364.85 {\tiny $\pm$ 9.97} & 534.04 {\tiny $\pm$ 9.79} & 478.46 {\tiny $\pm$ 6.99} & 455.94 {\tiny $\pm$ 4.81} & 512.91 {\tiny $\pm$ 4.46} \\
\midrule
Qwen3-VL-8B-Instruct (base) & 513.57 {\tiny $\pm$ 21.08} & 129.14 {\tiny $\pm$ 9.94} & 274.20 {\tiny $\pm$ 9.49} & 238.92 {\tiny $\pm$ 5.31} & 195.33 {\tiny $\pm$ 4.50} & 270.23 {\tiny $\pm$ 5.22} \\
\recipe{}-SFT (SFT on base) & 479.47 {\tiny $\pm$ 16.92} & 90.92 {\tiny $\pm$ 7.55} & 300.76 {\tiny $\pm$ 11.21} & 245.01 {\tiny $\pm$ 5.69} & 192.69 {\tiny $\pm$ 4.55} & 261.77 {\tiny $\pm$ 4.57} \\
\recipe{}-Zero  (RL on base) & \underline{1545.50 {\tiny $\pm$ 35.34}} & \underline{1222.69 {\tiny $\pm$ 21.52}} & \underline{1551.57 {\tiny $\pm$ 30.68}} & \underline{1262.18 {\tiny $\pm$ 39.95}} &  \underline{1192.71 {\tiny $\pm$ 20.74 }} & \underline{1354.93 {\tiny $\pm$ 13.68}} \\
\recipe{}  (RL on SFT) & \textbf{1644.43 {\tiny $\pm$ 17.53}} & \textbf{1430.88 {\tiny $\pm$ 22.00}} & \textbf{1603.36 {\tiny $\pm$ 18.62}} & \textbf{1352.30 {\tiny $\pm$ 14.14}} & \textbf{1528.51 {\tiny $\pm$ 25.75}} & \textbf{1511.90 {\tiny $\pm$ 8.95}} \\
\midrule
Maximum & 2351 & 2190 & 2336 & 2510 & 2191 & 2315.6 \\
\bottomrule
\end{tabular}}
\end{center}
\caption{Comparisons of \recipe{} with frontier models across five levels used for RL training. We measure level progress, defined as the $x$-axis distance traveled by Mario from the start of each level. Results are aggregated over runs and reported as mean $\pm$ standard error. The last row indicates the maximum achievable progress on each level.}
\label{tbl:first_five_levels}
\vspace{-0.1in}
\end{table}

\section{The Effectiveness of \recipe{}}\label{sec:recipe_exp}
In the following, we demonstrate the effectiveness of \recipe{}. All the training are conducted with Qwen3-VL-8B-Instruct \citep{bai2025qwen3} as the base model and the RL training is performed on the first five levels of the game. Detailed configurations are listed in \cref{app:recipe}.

\subsection{Superior Training Performances}
In \cref{tbl:first_five_levels}, we first report the performance of flagship frontier VLMs, including both proprietary and open-source models, on the five training levels used for RL training in \recipe{}, which exhibit very limited capability in the considered game environment.

We report the final performance of \recipe{} in \cref{tbl:first_five_levels}, together with two ablations: \recipe{}-SFT, initialized via supervised fine-tuning, and \recipe{}-Zero, trained with RL directly from the base model without SFT initialization. After training, \recipe{} achieves substantial improvements, typically tripling—and in many cases increasing by an order of magnitude—the average level progress compared to the base model, while also significantly outperforming frontier models. In particular, in terms of average level progress, \recipe{} improves over the base model by $5.59\times$ and over the best-performing frontier model (GLM-4.6V) by $2.95\times$. These results highlight the effectiveness of \recipe{}.

Furthermore, although \recipe{}-SFT alone does not provide observable gains over the base model, training on top of this initialization, i.e., \recipe{}, consistently yields better performance than \recipe{}-Zero across all levels. This observation supports the useful role of SFT initialization in enabling more effective RL training.

\subsection{Generalizations in Games}\label{sec:ood_result}

We evaluate the generalization capabilities of \recipe{} under three progressively challenging settings. First, we consider \textbf{in-game off-policy} evaluation, where the agent is tested on 50 manually curated states from the five training levels (i.e., 10 states each level). Although from the same training levels, they are sampled independently of the agent’s trajectories and thus induce a state distribution shift. Second, we assess \textbf{in-game generalization} by evaluating the agent on 50 manually collected states from the remaining five levels of \textit{Super Mario Land} (i.e., 10 states each level), which are entirely unseen during RL training but share the same game mechanics and visual structure. Finally, we examine \textbf{cross-game generalization} by testing the agent on all 32 levels of another game \textit{Super Mario Bros.}, representing a more substantial domain shift in terms of level design and visual appearance. Together, these settings provide a comprehensive evaluation of robustness to distribution shifts, ranging from off-policy state variations to unseen levels and entirely new games. Details of the evaluation setup are provided in \cref{appsub:evaluation}.

\begin{figure}[t]
\centering
\hfill
\begin{subfigure}{0.49\linewidth}
\centering
\includegraphics[width=\linewidth]{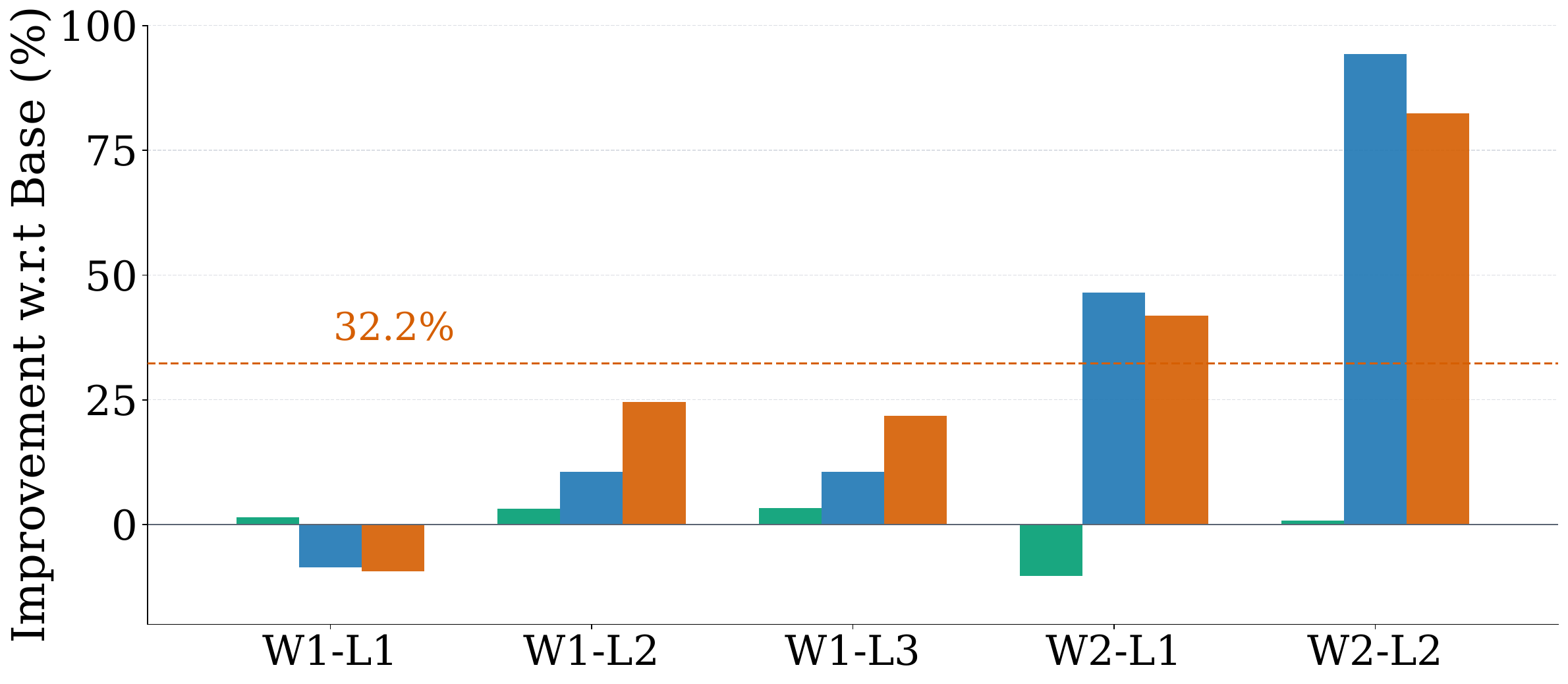}
\caption{In-game off-policy states (10 states/level).}
\label{fig:in_game_off_policy}
\end{subfigure}
\hfill
\begin{subfigure}{0.49\linewidth}
\centering
\includegraphics[width=\linewidth]{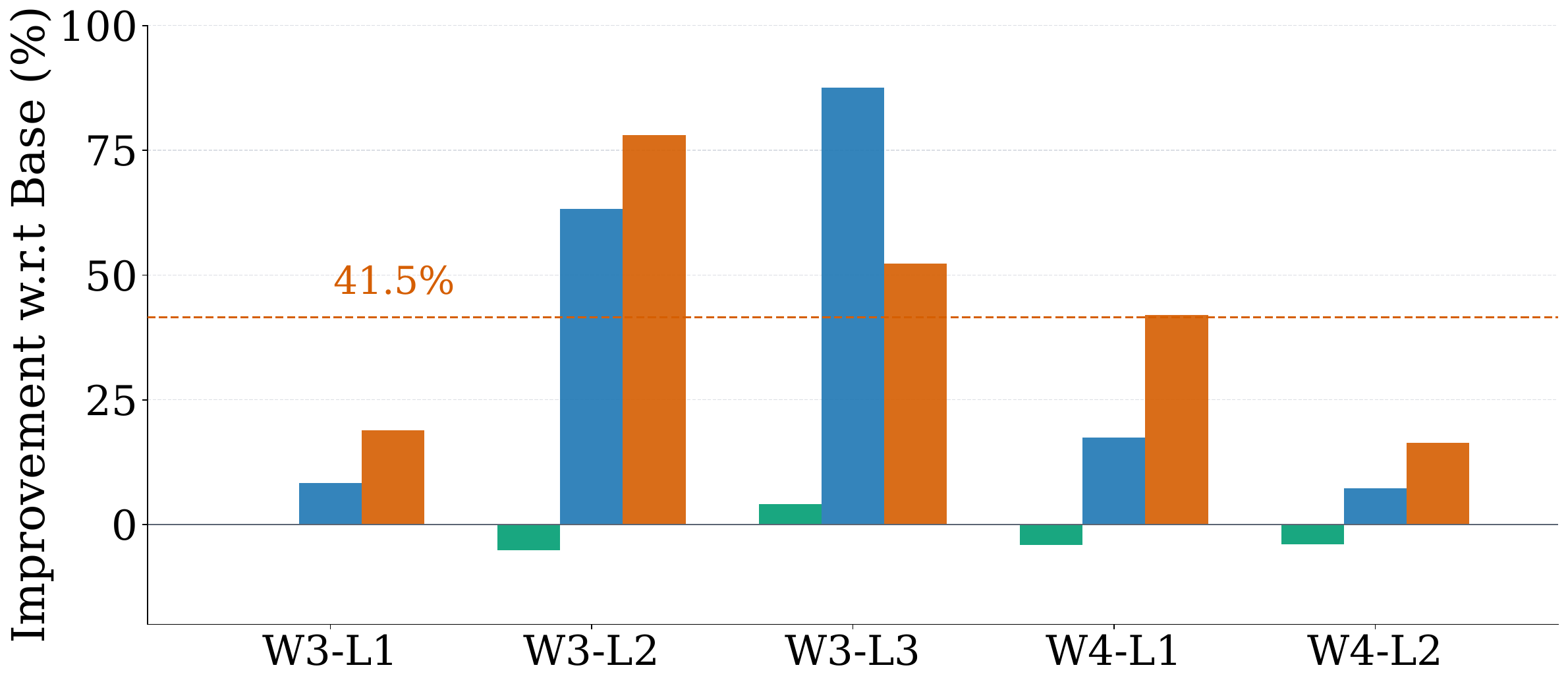}
\caption{In-game unseen states (10 states/level).}
\label{fig:in_game_unseen}
\end{subfigure}
\hfill
\begin{subfigure}{\linewidth}
\centering
\includegraphics[width=\linewidth]{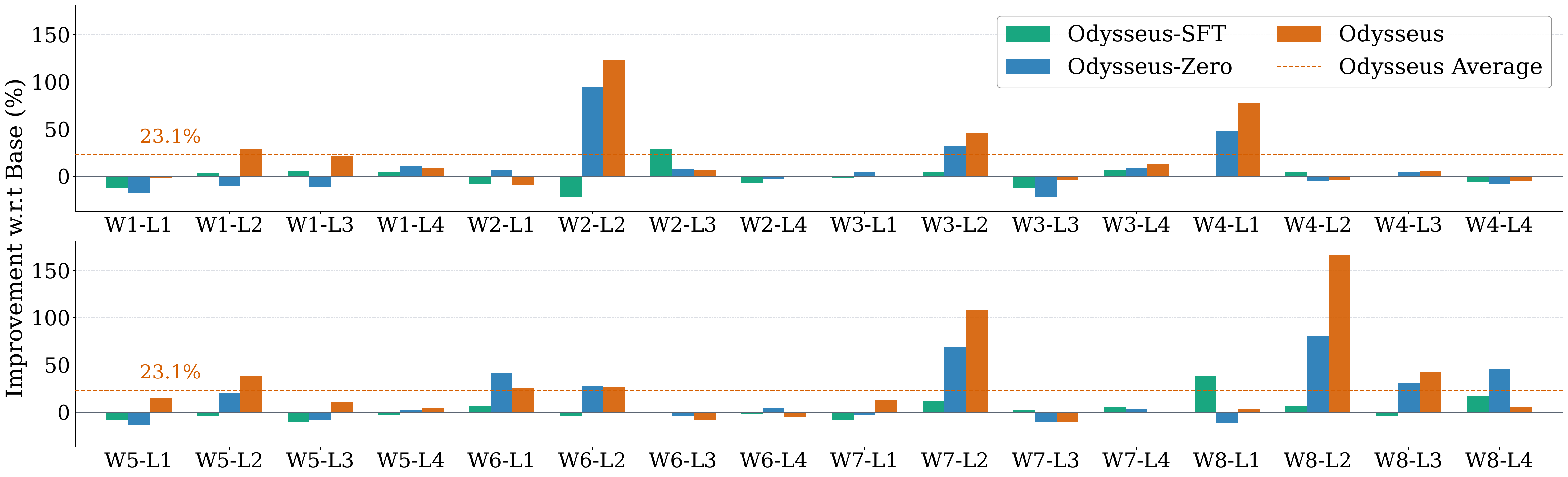}
\caption{Unseen game \textit{Super Mario Bros.} (32 levels)}
\label{fig:cross_game}
\end{subfigure}
\caption{Evaluation of \recipe{} under three generalization settings: in-game off-policy states (a), consisting of 10 manually collected states per level from the five training levels of \textit{Super Mario Land}; in-game unseen states (b), consisting of 10 manually collected states per level from the five held-out levels; and cross-game scenarios (c), spanning all 32 levels of \textit{Super Mario Bros}. Improvement is measured relative to the base pretrained VLM (before SFT or RL). For (a) and (b), the histograms report the average performance over the 10 states for each level, while for (c), performance is reported individually from the start of each level. The dotted horizontal line indicates the average improvement of \recipe{} over the base model across the scenarios in the corresponding subplot. Consistent gains across all settings demonstrate the strong generalization capability of \recipe{}.}
\vspace{-0.2in} 
\end{figure}

The relative improvements of the \recipe{}-series models on the first two in-game settings are reported in \cref{fig:in_game_off_policy} and \cref{fig:in_game_unseen}, where \recipe{} achieves average improvements of $32.2\%$ and $41.5\%$, respectively. The full results for the cross-game evaluation are presented in \cref{fig:cross_game}, where \recipe{} yields a relative improvement of $23.1\%$ over the base model on average. Overall, these results demonstrate that, despite being trained on only five levels of \textit{Super Mario Land}, \recipe{} exhibits clear signs of both in-game and cross-game generalization, highlighting the potential of this framework in building general agents.

\begin{table}[thb]
\small
\centering
    \begin{tabular}{c|ccc}
    \toprule
        Model & MMMUval & MathVision & RealWorldQA \\
        \midrule
        Qwen3-VL-8B-Instruct (base) & 69.00 & 54.64 & 71.11 \\
        \recipe{}-SFT (SFT on base) & 70.44 & 55.00 & 71.37 \\
        \recipe{}-Zero (RL on base)  & 70.22 & 54.44 & 70.72 \\
        \recipe{} (RL on SFT) & 70.77 & 53.52 & 71.11 \\
        \bottomrule
    \end{tabular}
    \caption{\recipe{}-series models maintain the base model's strong capabilities on general-purpose multi-modal benchmarks, in additional to the improved game performance.}
    \label{tbl:general_benchmark}
\end{table}

\subsection{Performances in General Domains}
Finally, given the extensive training in the game environment (over tens of millions of interaction samples), a natural concern is that the model may overfit and lose its general capabilities. To assess this, we evaluate the \recipe{}-series models on a set of general multi-modal benchmarks, including STEM-oriented tasks (MMMU~\citep{yue2024mmmu} and MathVision~\citep{wang2024measuring}) and real-world reasoning tasks (RealWorldQA~\citep{realworldqa}). The results, reported in \cref{tbl:general_benchmark}, show that compared to the base model, the \recipe{}-series models retain comparable performances. This suggests that \recipe{} can effectively inject new decision-making capabilities without compromising the model’s general-purpose strengths, highlighting its potential as a foundation for general-purpose agents.

\section{Conclusions}
In this work, we study the problem of training VLMs for long-horizon decision-making tasks, using the video game \textit{Super Mario Land} as a testbed, which requires 100+ interaction turns per episode. We first introduce an adapted PPO algorithm with a lightweight turn-level critic, which substantially improves training stability and sample efficiency over critic-free methods. Building on this, we show that fine-tuning pretrained VLMs via RL is significantly more sample-efficient than training agents from scratch with classical deep RL, highlighting the value of the encoded knowledge priors. Leveraging these insights, we propose \recipe{}, an open training framework for practical VLM agents that combines SFT initialization with multi-task RL training. \recipe{} enables stable training across multiple levels of the game simultaneously, yielding substantial performance gains over the base model and outperforming frontier models by large margins. Moreover, the resulting agents exhibit emergent generalization to both in-domain and out-of-domain settings, while retaining strong general-purpose capabilities. Overall, our results demonstrate that with an appropriate recipe, RL can be effectively scaled to long-horizon decision-making tasks for VLMs, providing a promising path toward more capable embodied agents.

\newpage
\section*{Acknowledgment}
The authors thank Princeton Language and Intelligence (PLI) and Princeton AI Lab for their support of this work, including computational resources and API credits, as well as many members of these groups for helpful discussions and suggestions. CJ acknowledges the support from NSF-OAC-2411299, NSF-IIS-2239297, and Sloan Research Fellowship.

\bibliography{colm2026_conference}
\bibliographystyle{colm2026_conference}

\newpage
\appendix

\section{Extended Related Work}
\label{app:extended_related_work}
\paragraph{Games and Simulated Environments.} 
Games and simulated environments have long served as controlled testbeds during the development of modern machine learning. In classical deep RL, benchmarks such as ALE~\citep{bellemare2013arcade} and MuJoCo~\citep{duan2016benchmarking} played a central role in studying algorithms learning from interactions with the environments. Later benchmarks extend this paradigm to more complex and advanced settings, including multi-agent RL~\citep{mordatch2017emergence,berner2019dota,vinyals2019grandmaster,li2024fightladder}, multi-task generalization~\citep{nichol2018retro,yu2020meta}, embodied tasks~\citep{ALFRED20,ALFWorld20,liu2023libero}, and open-world environments~\citep{fan2022minedojo,tan2025lumine}. 
More recently, several works have begun to use video games as direct testbeds for foundation models~\citep{hu2025lmgame,karten2025pokeagent,zhang2025videogamebench,park2025orak,foundation2026arc}. However, the recent exploration of RL for VLMs in simulated environments is mostly focused on short-horizon scenarios, such as AlfWorld~\citep{ALFWorld20}, Sokoban, and FrozenLake~\citep{wang2025vagen}. In contrast to these settings, we focus on \textit{Super Mario Land} as a compact but appealing testbed for VLMs in long-horizon embodied control: it imposes substantially richer spatial grounding and closed-loop control than short-horizon gridworld-style tasks, while remaining lightweight and easy to scale for controlled studies compared with large open-world simulators.

\paragraph{Foundation Models for Decision-Making.}
The integration of foundation models into sequential decision-making has evolved through several distinct paradigms. Initial efforts treated control primarily as a sequence modeling problem, training Transformers from scratch on offline trajectory data~\citep{janner2021offline,chen2021decision,li2023survey}. This approach demonstrated strong potential for task generalization~\citep{lee2022multi,reed2022generalist} and skill composition in open-ended environments~\citep{baker2022video,fan2022minedojo}. As large language and vision-language models advanced, a second line of work emerged that leverages frozen large models with context engineered states and agentic scaffolding or harnesses in long context sequential decision-making tasks, such as navigating complex video games like Pokémon~\citep{karten2025pokeagent,karten2025pokechamp,comanici2025gemini} and Minecraft~\citep{wang2023voyager}, or facilitating robotics manipulation tasks~\citep{huang2023voxposer,saycan2022arxiv}. More recently, the field has shifted toward fine-tuning pretrained foundation models directly for embodied control. This has yielded highly capable agents in both robotic manipulation~\citep{black2024pi_0,liu2024rdt,team2025gemini} and cross-game generalization~\citep{tan2025lumine,bolton2025sima,wang2025game}. While highly promising, these fine-tuning approaches heavily depend on SFT with large amounts of high-quality, action-labeled demonstration data. In contrast, our work focuses on the RL perspective, investigating how to effectively adapt foundation models for long-horizon decision-making tasks without relying on extensive SFT data.

\paragraph{RL from Classical Control to Foundation-Model Agents.}
RL has a long history in training neural networks for sequential decision-making, spanning policy-gradient~\citep{williams1992simple,silver2014deterministic}, value-based~\citep{mnih2015human,van2016deep} and actor-critic~\citep{schulman2017proximal,haarnoja2018soft} methods. Particularly, sample efficiency is a key challenge for RL algorithms as they require far more environmental interactions than humans during training time, motivating a line of research on sample-efficient RL~\citep{deisenroth2011pilco,kaiser2019model,hafner2019dream,schwarzer2020data,ye2021mastering}. More recently, RL has also become a central ingredient for improving foundation models, especially in reasoning-oriented settings with verifiable rewards, through REINFORCE-style and critic-free methods~\citep{ahmadian2024back,hu2025reinforce++,shao2024deepseekmath}. A growing body of work further studies RL for multi-turn language and vision-language agents~\citep{zhai2024fine,chen2025era,wang2025ragen,wang2025vagen,feng2025group,li2026salt,feng2025group,he2026hierarchy}, often introducing specialized machinery for trajectory decomposition, turn-level and token-level advantage estimation, or hierarchical credit assignment, and evaluating on environments with relatively short horizons (20--30 turns). In contrast to these works, we focus specifically on long-horizon, visually grounded embodied environments that require 100+ turns of interaction with chain-of-thought reasoning, and through rigorous ablations, we show that a comparatively simple PPO-based approach with the right critic design is sufficient to make RL stable and effective.

\section{Details of the Interaction Protocol}\label{app:interaction}
First, the full prompt used for instructing the VLM agent to interact with the game environment is provided in the following.

\begin{tcolorbox}[title=Prompt for the Agent]
\small
You are playing Super Mario Land.\\

The goal is to progress through levels, collect coins and power-ups when safe, and ultimately finish the game by rescuing Princess Daisy.\\

You can control the game by pressing buttons on the Game Boy.\\

Available buttons:\\
- 'a': Jump (used to make Mario jump)\\
- 'b': Run/Shoot (hold to run faster or shoot fireballs if available)\\
- 'up': Climb ladders or vines (if present)\\
- 'down': Crouch or enter pipes (when standing on a pipe)\\
- 'left': Move Mario left\\
- 'right': Move Mario right\\
- 'noop': Do nothing (used to wait for a brief period without performing any action)\\

Please analyze the game screen and decide which buttons to press to progress.\\

Return your answer as follows:\\
1. Button sequence: a list of buttons to press simultaneously\\
2. Each button should be one of: 'a', 'b', 'up', 'down', 'left', 'right', 'noop'\\

First describe what you see on the screen in <perception></perception>. Then, in <reasoning></reasoning>, break down your reasoning step by step, justifying each action you consider. Output your final action in <answer>['button1', 'button2', ...]</answer>.\\

The maximum number of buttons you can press simultaneously in one turn is 2.
\end{tcolorbox}

At each turn, the agent observes the current game frame and the instruction prompt, and produces an action in the prescribed format. Since the original game resolution of $160 \times 144$ is relatively low compared with the visual resolutions used during modern VLM pre-training, we up-sample the image by a factor of $8$ to $1280 \times 1152$. The final action is extracted from the \texttt{<answer></answer>} field and executed in the game environment as button presses. As described in \cref{sec:protocol}, we adopt a frame-skipping mechanism: when the action includes a jump (i.e., \texttt{a}), the environment advances for 15 frames; otherwise, it advances for 5 frames, including the case of \texttt{noop}, where the game proceeds without any button input.

\section{Details of RL Algorithms for VLM Training}\label{app:vlm_rl}
In this section, we provide the implementation details of the RL algorithms discussed in \cref{sec:rl_comparison}, including the proposed adapted version of PPO.

\subsection{Advantage Constructions}
Provided with a dataset batch $\mathcal{D}$ of trajectories collected via the policy from the previous training step, denoted as $\pi_{\text{old}}$, all methods are instantiated with the same surrogate loss \citep{schulman2017proximal}:
\begin{align*}
    \gL(\theta) = \E_{o_t,a_t\sim \mathcal{D}} \left[ \min\left( \frac{\pi_\theta(a_t\vert o_t)}{\pi_{\theta_\text{old}}(a_t\vert o_t)}\hat{A_t}, \text{clip}\left( \frac{\pi_\theta(a_t\vert o_t)}{\pi_{\theta_\text{old}}(a_t\vert o_t)}, 1-{\epsilon_{\text{low}}}, 1+{\epsilon_{\text{high}}} \right)\hat{A_t} \right)\right],
\end{align*}
where $\epsilon_{\text{low}}$ and $\epsilon_{\text{high}}$ are the clipping factors while $\hat{A_t}$ is the advantage estimator at turn $t$. The key differences of the algorithms are in how advantages are constructed, which are detailed in the following.

\paragraph{PPO with a Turn-level Critic.} We first discuss the adapted version of PPO proposed in this work. A \textit{turn-level} critic model $V_\phi(o_t)$ is learned to approximate the value of policy starting from $s_t$, and use the discounted return-to-go $\hat{R}_t=\sum_{i\geq t}\gamma^{i-t}r_i$ as the target value:
\begin{equation*}
    \mathcal{L}_{V}(\phi) = \E_{o_t\sim \mathcal{D}} \left[ \text{SmoothL}_1 \left( V_\phi(o_t) - \hat{R}_t \right) \right].
\end{equation*}

The raw per-turn advantage is first computed as 
\begin{equation*}
    \tilde{A}_t = \hat{R}_t - V_\phi(o_t).
\end{equation*}
We perform one more batch-level variance normalization over it to obtain the advantage as
\begin{equation*}
    \hat{A}_t = \frac{\tilde{A}_t}{\sigma\left(\{\tilde{A}_{t'}: o_{t'}\in \mathcal{D}\}\right)},
\end{equation*}
where $\sigma(\cdot)$ denotes standard deviation.

When performing positive-advantage filtering, we instead keep the signed advantages and normalize them over them:
\begin{equation}
    \hat{A}_t = \frac{\max(\tilde{A}_t,0)}{\sigma\left(\{\tilde{A}_{t'}: \tilde{A}_{t'}>0, o_{t'} \in \mathcal{D}\}\right)}.
\end{equation}

\paragraph{GRPO with Outcome Rewards.}
For GRPO with outcome rewards, all turns within the same trajectory share a common trajectory-level outcome return. Specifically, for a trajectory
\begin{equation*}
    \tau=(o_0,a_0,r_0,\cdots,o_{T-1},a_{T-1},r_{T-1}),
\end{equation*}
 we define the outcome return as the cumulative reward over the full trajectory:
\begin{equation}
    \hat{R}^{\text{out}}(\tau) = \sum_{i=0}^{T-1}r_i,
\end{equation}
which is equivalent to setting the discount factor $\gamma=1$ when calculating return-to-go. This same outcome return is assigned to every turn $t$ in the trajectory. Following standard GRPO \citep{shao2024deepseekmath}, we standardize the trajectory-level outcome return over the batch:
\begin{equation}
    \hat{A}_t = \frac{\hat{R}^{\text{out}}(\tau) - \E_{\tau'\in \mathcal{D}}[\hat{R}^{\text{out}}(\tau')]}{\sigma\left(\{\hat{R}^{\text{out}}(\tau'): \tau' \in \mathcal{D}\}\right)}.
\end{equation}

When performing positive advantage filtering, it is further set that $\hat{A}_t =\max(0, \hat{A}_t)$.

\paragraph{GRPO with Process Rewards.}
For GRPO with process rewards, following \citet{shao2024deepseekmath}, we first standardize turn rewards over the current batch before constructing the return-to-go as
\begin{equation*}
    \tilde{r}_{t} = \frac{r_t - \E_{r_{t'}\in \mathcal{D}}[r_{t'}]}{\sigma(\{r_{t'}:r_{t'}\in \mathcal{D}\})}.
\end{equation*}
The per-turn training signal as the un-discounted return-to-go of the standardized process rewards:
\begin{equation*}
    \hat{A}_t = \sum_{i=t}^{T-1}\tilde{r}_i.
\end{equation*} 
When performing positive advantage filtering, it is further set that $\hat{A}_t =\max(0, \hat{A}_t)$.

\paragraph{Reinforce++.}
For Reinforce++ \citep{hu2025reinforce++}, we use the same turn-level discounted return-to-go as PPO, i.e., $\hat{R}_t=\sum_{i\geq t}\gamma^{i-t}r_i$, and standardize it over the entire batch as the advantage as
\begin{equation*}
    \hat{A}_t = \frac{\hat{R}_t - \E_{o_{t'}\in \mathcal{D}}[\hat{R}_{t'}]}{\sigma\left(\{\hat{R}_{t'}:o_{t'}\in \mathcal{D}\}\right)}.
\end{equation*}

\subsection{Experimental Details and Additional Results for \cref{subsec:rl_alg_comp}}
\begin{figure}[t]
\centering
\hfill
\begin{subfigure}{0.49\linewidth}
\centering
\includegraphics[width=\linewidth]{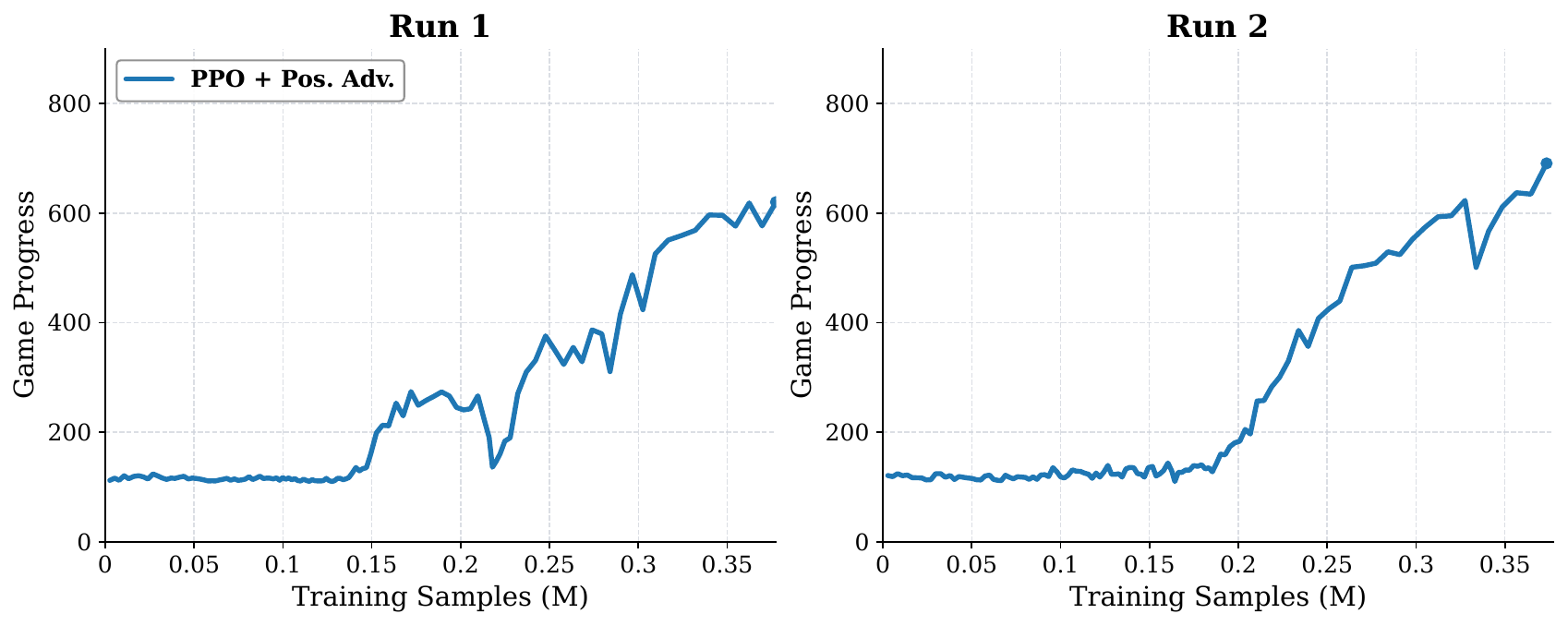}
\caption{PPO + Pos. Adv.}
\end{subfigure}
\hfill
\begin{subfigure}{0.49\linewidth}
\centering
\includegraphics[width=\linewidth]{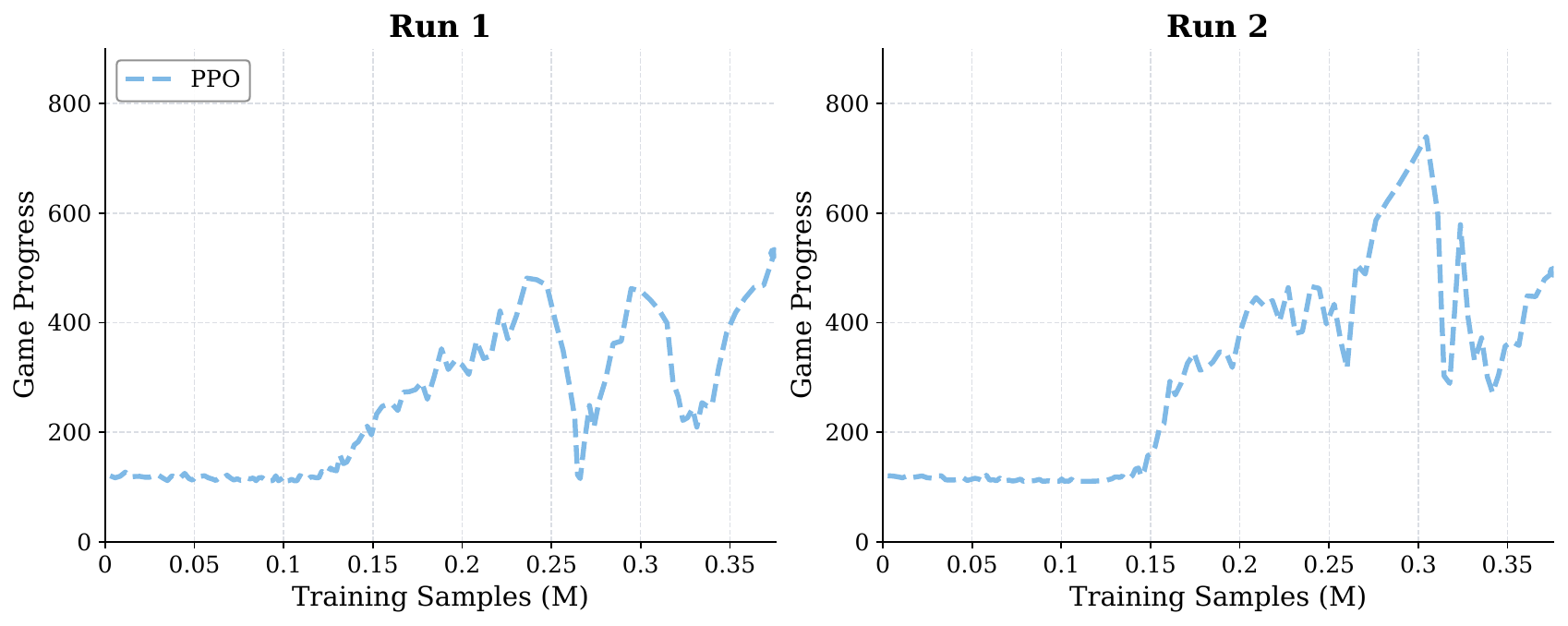}
\caption{PPO.}
\end{subfigure}
\hfill
\\
\hfill
\begin{subfigure}{0.49\linewidth}
\centering
\includegraphics[width=\linewidth]{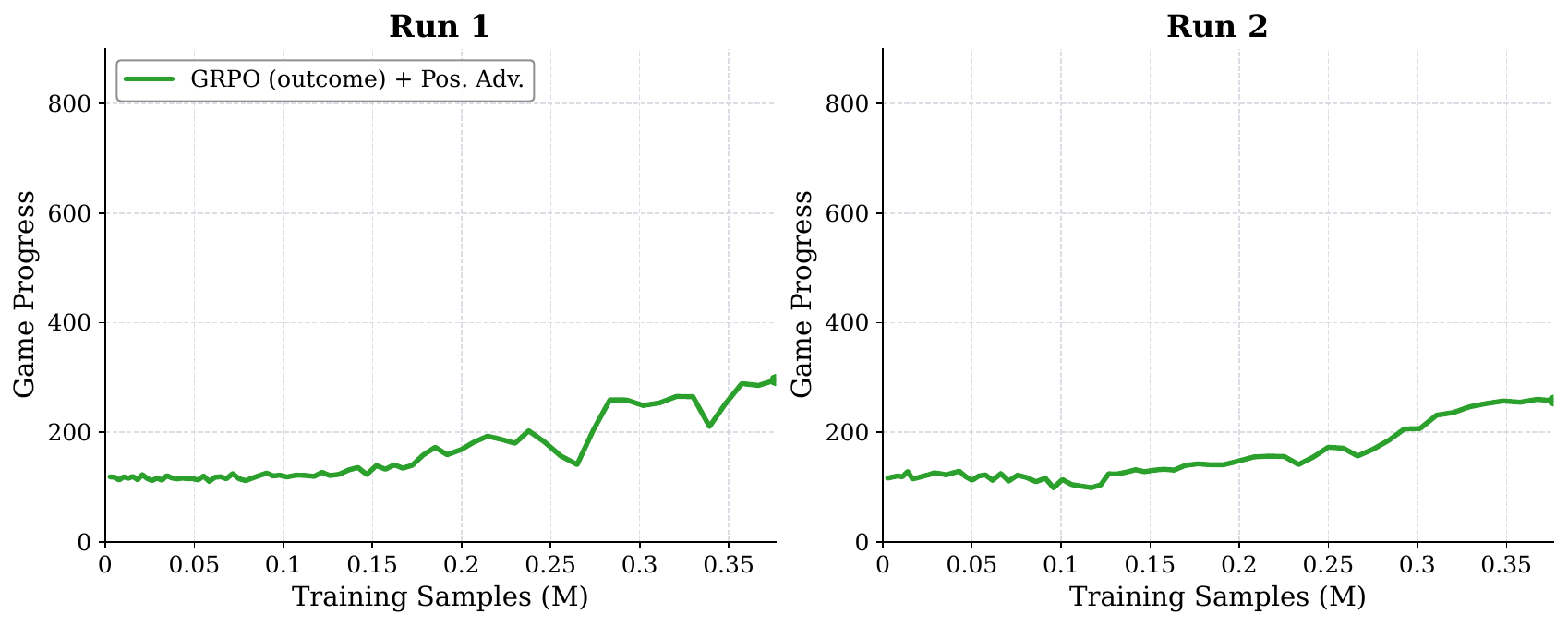}
\caption{GRPO (outcome) + Pos. Adv.}
\end{subfigure}
\hfill
\begin{subfigure}{0.49\linewidth}
\centering
\includegraphics[width=\linewidth]{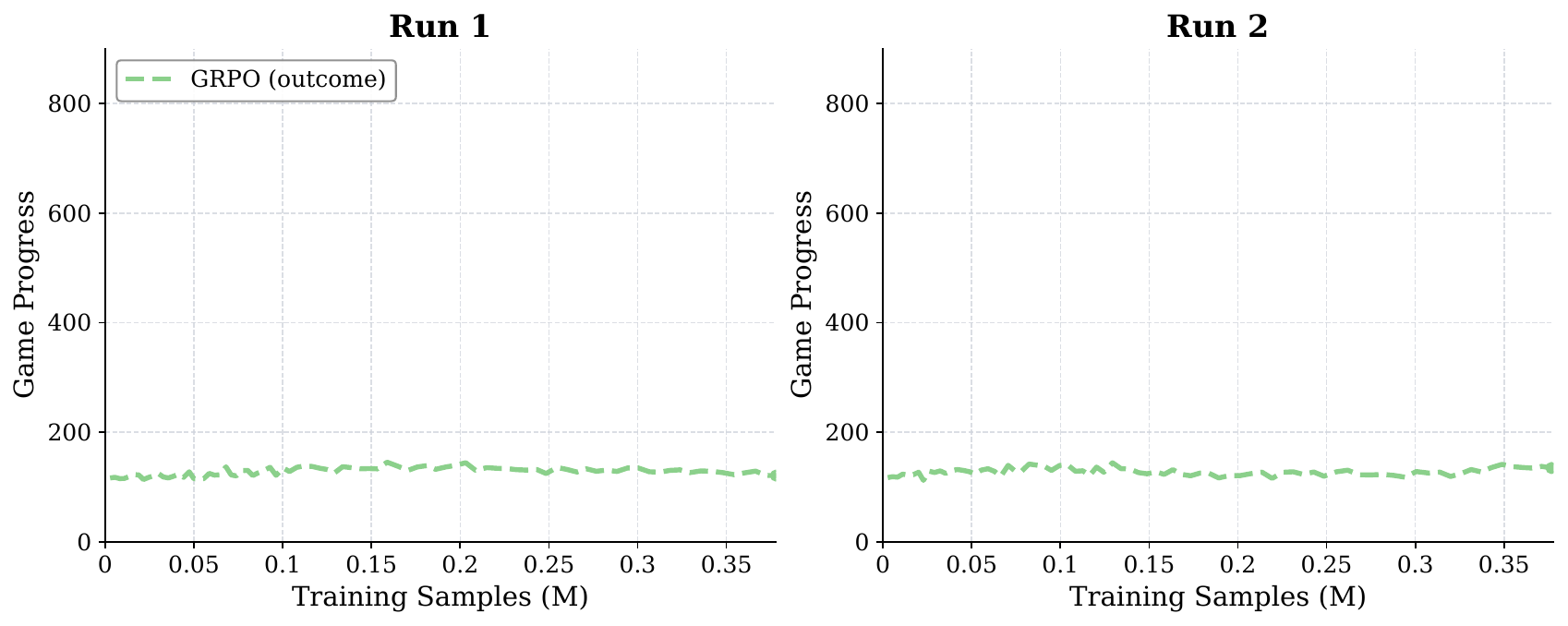}
\caption{GRPO (outcome).}
\end{subfigure}
\hfill
\\
\hfill
\begin{subfigure}{0.49\linewidth}
\centering
\includegraphics[width=\linewidth]{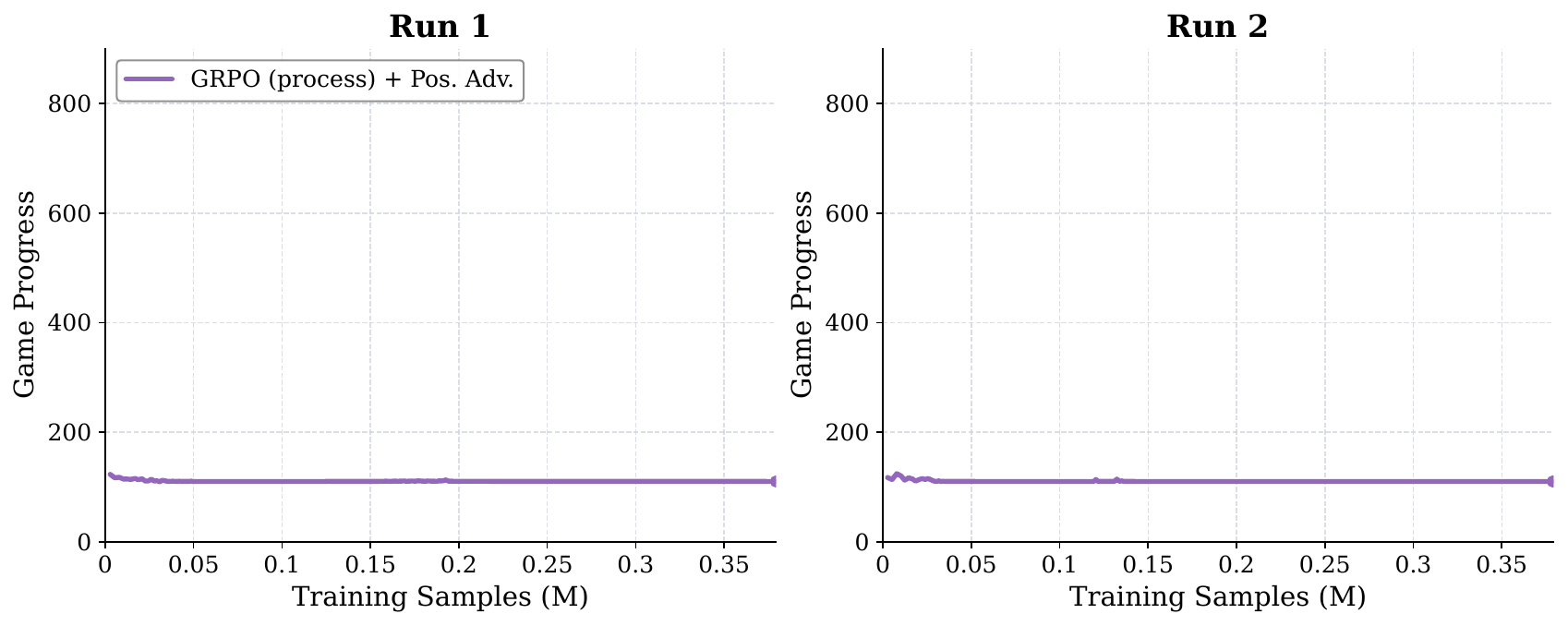}
\caption{GRPO (process) + Pos. Adv.}
\end{subfigure}
\hfill
\begin{subfigure}{0.49\linewidth}
\centering
\includegraphics[width=\linewidth]{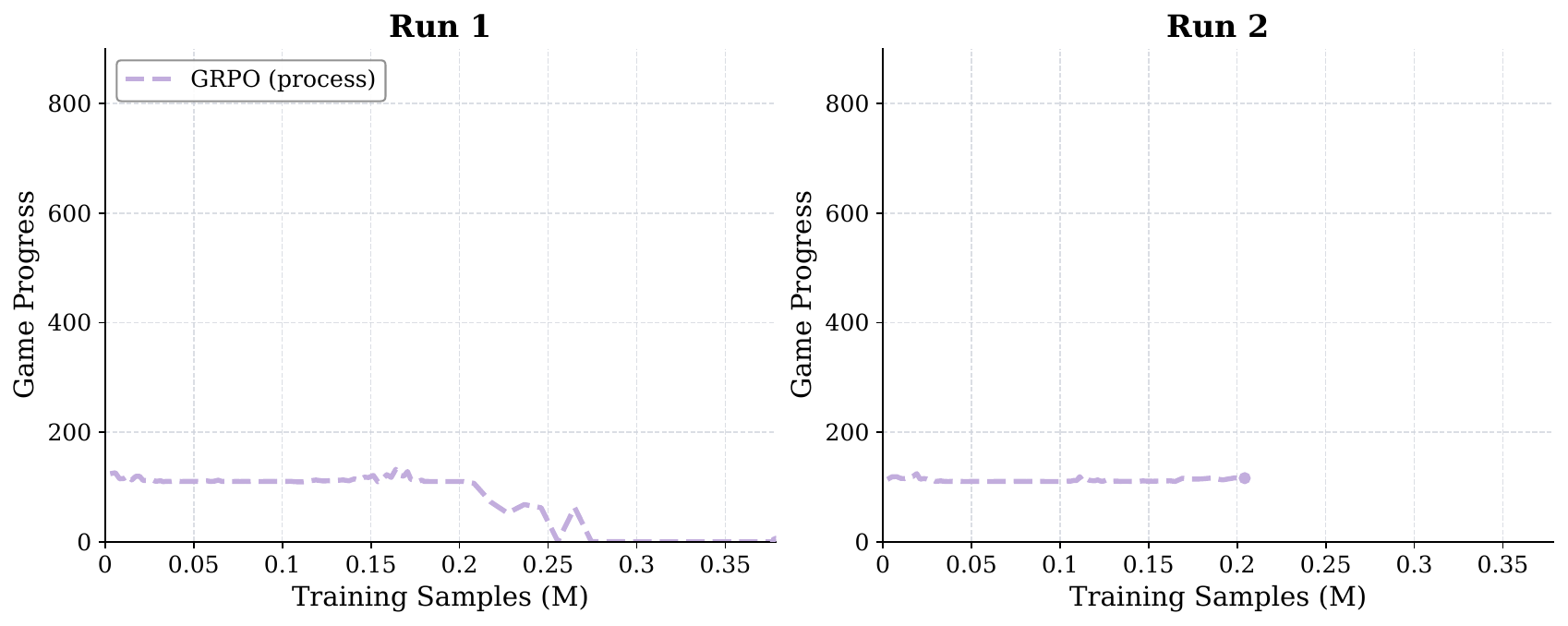}
\caption{GRPO (process).}
\end{subfigure}
\hfill
\begin{subfigure}{0.49\linewidth}
\centering
\includegraphics[width=\linewidth]{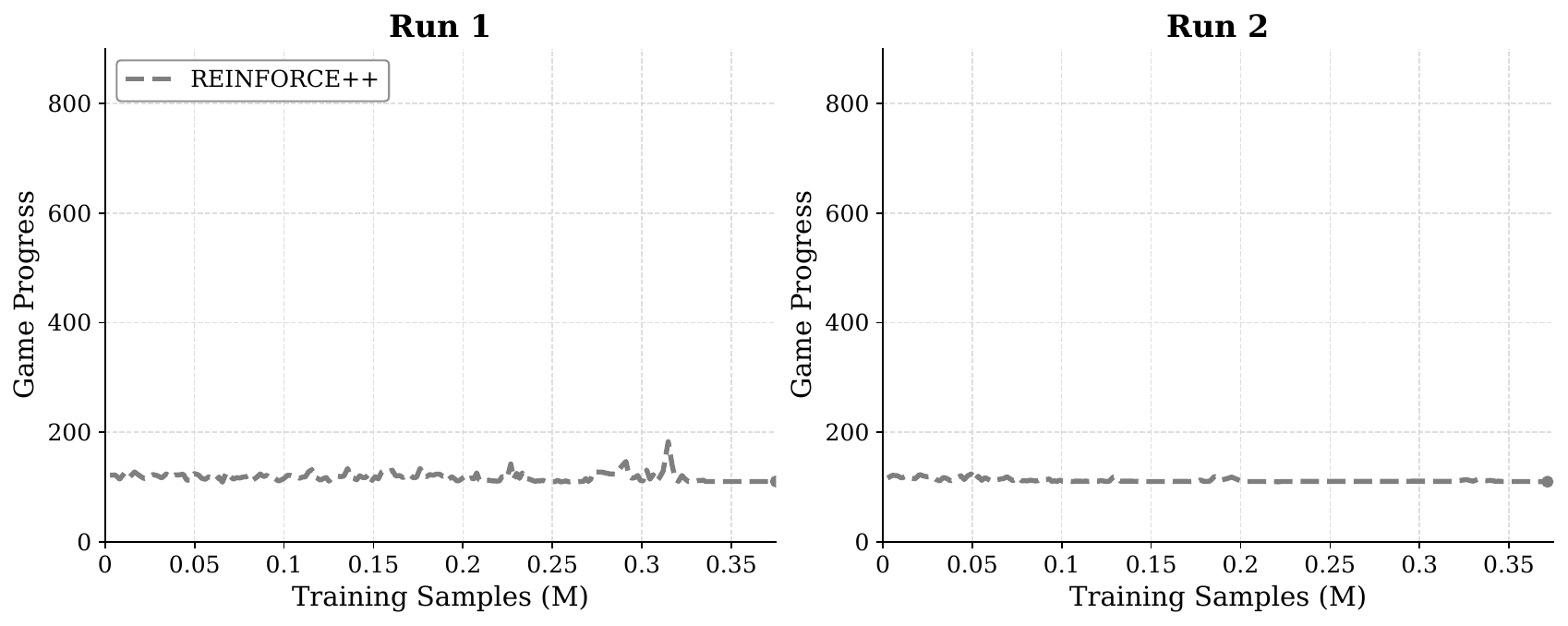}
\caption{Reinforce++.}
\end{subfigure}
\hfill
\caption{ Comparison of VLM-based RL training methods. Individual runs are plotted separately without smoothing.}
\label{fig:vlm_compare_full}
\end{figure}

In \cref{tbl:vlm_rl}, we provide the detailed configurations of experiments reported in \cref{subsec:rl_alg_comp}, which are shared across the tested algorithms. During all experiments, the training is performed on all components of the base model (i.e., Qwen3-VL-8B-Instruct), including vision encoder, multi-modal projector, and language model backbone. For the adapted PPO, the network architecture for the CNN-based critic and its training hyper-parameters are further detailed in \cref{app:recipe}, which are shared between experiments in \cref{subsec:rl_alg_comp} and \cref{sec:recipe_exp}. Additional experimental results, including all seven evaluated methods and their individual runs, are provided in \cref{fig:vlm_compare_full}.

\begin{table}[tb]
\centering
\small
\begin{tabular}{@{}ll@{}}
\toprule
Parameter & Value \\
\midrule
max turns & $80$ \\
top p & 1 \\
temperature & 1 \\
max response length & $1024$ \\
number of trajectories per batch  & $128$ \\
optimization epochs per training step & $1$ \\
mini-batch size & $1024$ \\
discount factor $\gamma$ & $0.95$ \\
policy clip range $\epsilon_\text{high}$,  $\epsilon_\text{low}$ & $0.28, 0.2$ \\
learning rate & $5\times 10^{-5}$ \\
learning rate scheduler & constant \\
\bottomrule
\end{tabular}
\caption{Hyper-parameters for comparisons of VLM-based RL algorithms.}
\label{tbl:vlm_rl}
\end{table}

\begin{figure}[t]
\centering
\begin{subfigure}{\linewidth}
\centering
\includegraphics[width=0.5\linewidth]{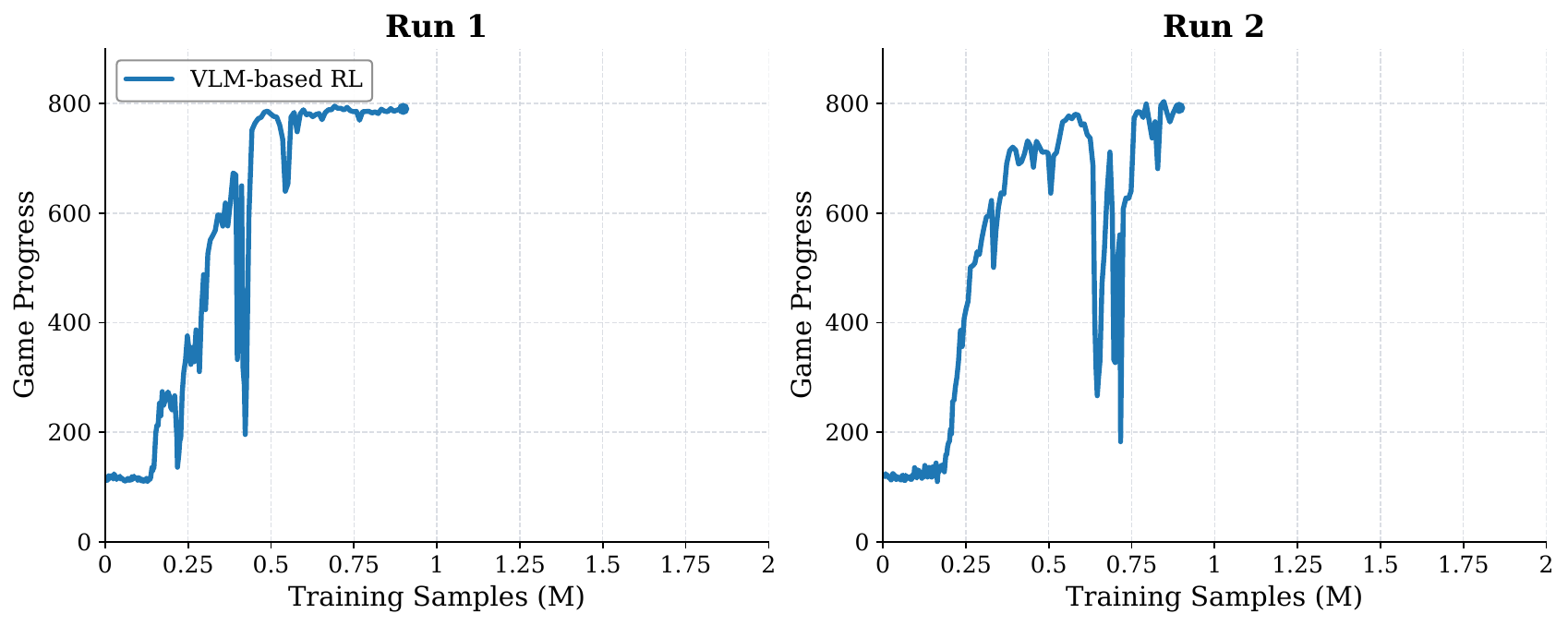}
\caption{VLM-based RL (i.e., PPO with a turn-level CNN critic and positive advantage filtering).}
\end{subfigure}
\\
\begin{subfigure}{\linewidth}
\centering
\includegraphics[width=0.75\linewidth]{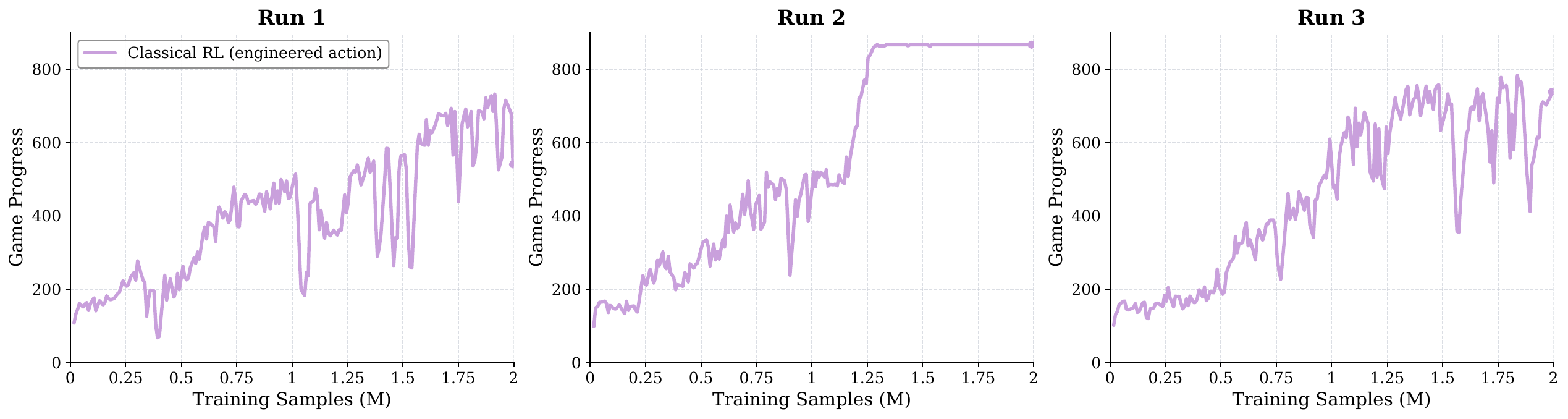}
\caption{Classical RL (i.e., PPO training a CNN policy from scratch) using the engineered action space.}
\end{subfigure}
\begin{subfigure}{\linewidth}
\centering
\includegraphics[width=0.75\linewidth]{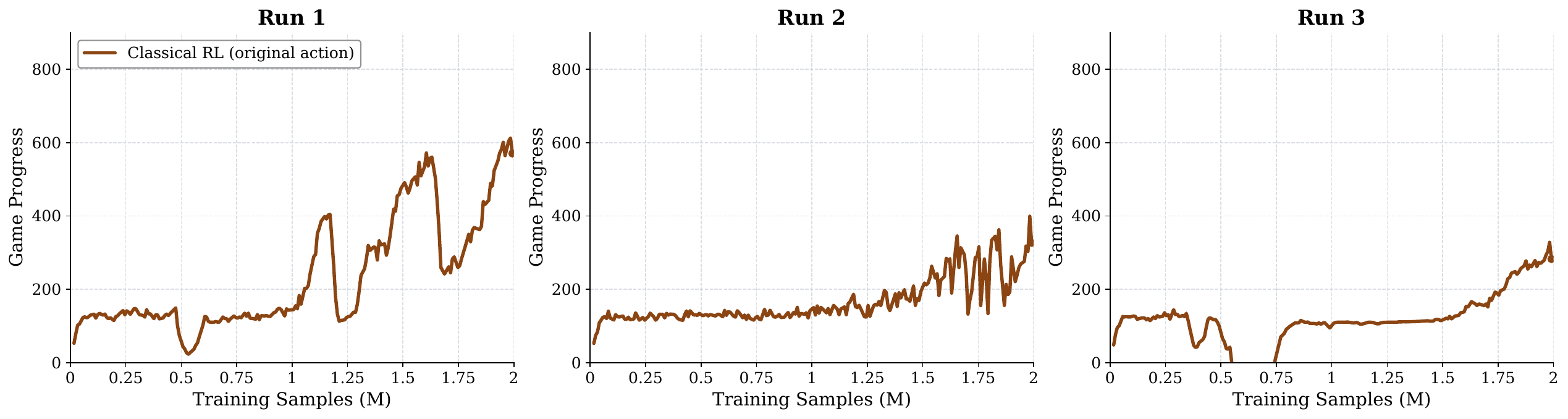}
\caption{Classical RL (i.e., PPO training a CNN policy from scratch) using the original action space.}
\end{subfigure}
\caption{Comparison between VLM-based RL and classical RL. Individual runs are plotted separately without smoothing.}
\label{fig:classical_rl_compare_full}
\end{figure}

\section{Details of Comparisons with Classical Deep RL}\label{app:classical_rl}
In this section, we provide the implementation details of the classical deep RL baselines used in \cref{sec:classical_rl}, with the complete experimental results provided in \cref{fig:classical_rl_compare_full}.

\paragraph{Environment and Action Spaces.}
The environment used for classical deep RL training is set to be exactly the same as that used in VLM training to ensure comparable results, including but not limited to, rewards and frame-skipping mechanisms. The major differences lie in the action spaces. As mentioned in \cref{sec:protocol}, VLMs are allowed to output up to press two buttons at the same time out from a total list of seven (i.e., \texttt{a}, \texttt{b}, \texttt{up}, \texttt{down}, \texttt{left}, \texttt{right}, \texttt{noop}). For classical RL, the ``original'' action space (with a size of 22) contains \texttt{noop} and up to two buttons from the remaining list. The engineered action space is more tailored towards human game-playing behaviors, which is illustrated in \cref{tbl:engineered_action_space}.

\begin{table}[tb]
\centering
\small
\begin{tabular}{@{}ll@{}}
\toprule
Action & Buttons \\
\midrule
\texttt{RIGHT} & \{\texttt{right}\} \\
\texttt{RIGHT\_JUMP} & \{\texttt{right}, \texttt{a}\} \\
\texttt{RIGHT\_SPRINT\_JUMP} & \{\texttt{right}, \texttt{a}, \texttt{b}\} \\
\texttt{LEFT} & \{\texttt{left}\} \\
\texttt{LEFT\_JUMP} & \{\texttt{left}, \texttt{a}\} \\
\texttt{RIGHT\_SPRINT} & \{\texttt{right}, \texttt{b}\} \\
\texttt{LEFT\_SPRINT} & \{\texttt{left}, \texttt{b}\} \\
\texttt{JUMP} & \{\texttt{a}\} \\
\bottomrule
\end{tabular}
\caption{The engineered action space used in the classical RL baselines.}
\label{tbl:engineered_action_space}
\end{table}

\paragraph{Algorithm and hyper-parameters.}
We use the PPO with a CNN policy implemented in Stable-Baselines 3 (SB3)~\citep{stable-baselines3} as the classical RL baseline. The policy is instantiated as \texttt{CnnPolicy} with the default SB3 \texttt{NatureCNN} backbone, with hyper-parameters listed in \cref{tbl:sb3}. Note that the chosen learning rate is selected via grid-search over $[5 \times 10^{-5},1 \times 10^{-4},1.5 \times 10^{-4},2.5 \times 10^{-4}]$. Training is run for up to $2 \times 10^6$ samples, i.e., state-action pairs, and evaluations are performed every $10^4$ samples for $128$ trajectories with different random seeds. 

\begin{table}[h]
\centering
\small
\begin{tabular}{@{}ll@{}}
\toprule
Parameter & Value \\
\midrule
batch size & $8192$ \\
optimization epochs & $4$ \\
mini-batch size & $1024$ \\
learning rate & $5\times 10^{-5}$ \\
discount factor $\gamma$ & $0.95$ \\
GAE parameter $\lambda$ & $0.95$ \\
policy clip range $\epsilon$ & $0.1$ \\
value clip range & $0.1$ \\
maximum gradient norm & $0.5$ \\
target KL & $0.03$ \\
\bottomrule
\end{tabular}
\caption{Hyper-parameters for classical RL experiments with PPO via SB3.}
\label{tbl:sb3}
\end{table}

\section{Details of Training and Evaluations of \recipe{}}\label{app:recipe}
\subsection{Details of the CNN Critic}
We use the same CNN backbone as~\citep{mnih2015human} for the critic learning and provide the detailed hyper-parameters for critic learning in \cref{tbl:critic}.

\begin{table}[h]
\centering
\small
\begin{tabular}{@{}ll@{}}
\toprule
Parameter & Value \\
\midrule
hidden dimension & $512$ \\
optimization epochs per training step & $1$ \\
weight decay & $0.0$ \\
gradient clip & $1.0$ \\
learning rate & $3\times 10^{-4}$ \\
learning rate scheduler & constant \\
\bottomrule
\end{tabular}
\caption{Hyper-parameters for CNN critic learning in \recipe{}.}
\label{tbl:critic}
\end{table}

\subsection{Other Training Details}
During all experiments, the training is performed on all components of the base model (i.e., Qwen3-VL-8B-Instruct), including vision encoder, multi-modal projector, and language model backbone. The SFT initialization is performed using the Qwen3-VL repo\footnote{\url{https://github.com/QwenLM/Qwen3-VL}} with configurations listed in \cref{tbl:recipe_sft}. RL training is conducted under a substantially modified version of the VeRL framework\footnote{\url{https://github.com/verl-project/verl}} \citep{sheng2024hybridflow} with configurations listed in \cref{tbl:recipe_rl}.

\subsection{Evaluation Details}\label{appsub:evaluation}

We compute mean and standard error over multiple runs of each model. For close-source proprietary models requiring API access, we compute summary statistics over 8 runs. For remaining open-source models, we compute summary statistics over 256 runs. We use the same set of hyperparameters in inference as in training.

For the in-game generalization evaluations, we manually collect game states across 10 levels in \textit{Super Mario Land}. While for cross-game generalization evaluations, the game environment initializes from start in 32 levels of \textit{Super Mario Bros.}

\begin{table}[h]
\centering
\small
\begin{tabular}{ll}
\toprule
Parameter & Value \\
\midrule
dataset size & 5058 \\
total epoch & 1 \\
batch size & 128 \\
gradient accumulation step & 2 \\
learning rate & $1\times10^{-7}$ \\
learning rate scheduler & cosine \\
warmup ratio & 0.03 \\

\bottomrule
\end{tabular}
\caption{Hyper-parameters for SFT training in \recipe{}.}
\label{tbl:recipe_sft}
\end{table}

\begin{table}[h]
\centering
\small
\begin{tabular}{ll}
\toprule
Parameter & Value \\
\midrule
training steps of \recipe{} & 190 \\
training steps of \recipe{}-Zero & 175 \\
max turns & $400$ \\
top p & 1 \\
temperature & 1 \\
max response length & $4096$ \\
number of trajectories per batch  & $1024$ \\
optimization epochs per training step & $1$ \\
mini-batch size & $4096$ \\
discount factor $\gamma$ & $0.95$ \\
policy clip range $\epsilon_\text{high}$,  $\epsilon_\text{low}$ & $0.28, 0.2$ \\
learning rate & $1\times 10^{-6}$ \\
learning rate scheduler & constant \\
\bottomrule
\end{tabular}
\caption{Hyper-parameters for RL training in \recipe{}.}
\label{tbl:recipe_rl}
\end{table}

\section{Additional Visualizations}
\begin{figure}
    \centering
    \begin{subfigure}[t]{0.23\textwidth}
        \centering
        \includegraphics[width=\linewidth]{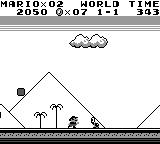}
    \end{subfigure}%
    \hfill
    \begin{subfigure}[t]{0.23\textwidth}
        \centering
        \includegraphics[width=\linewidth]{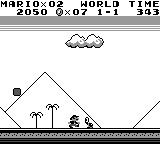}
    \end{subfigure}%
    \hfill 
    \begin{subfigure}[t]{0.23\textwidth}
        \centering
        \includegraphics[width=\linewidth]{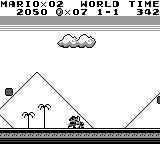}
    \end{subfigure}%
    \hfill 
    \begin{subfigure}[t]{0.23\textwidth}
        \centering
        \includegraphics[width=\linewidth]{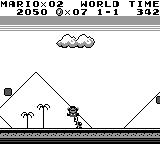}
    \end{subfigure}

    \vspace{1em}

    \begin{subfigure}[t]{0.23\textwidth}
        \centering
        \includegraphics[width=\linewidth]{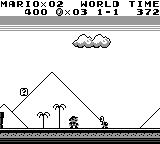}
    \end{subfigure}%
    \hfill 
    \begin{subfigure}[t]{0.23\textwidth}
        \centering
        \includegraphics[width=\linewidth]{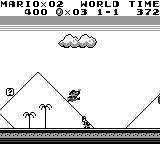}
    \end{subfigure}%
    \hfill 
    \begin{subfigure}[t]{0.23\textwidth}
        \centering
        \includegraphics[width=\linewidth]{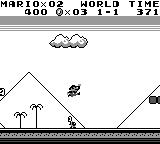}
    \end{subfigure}%
    \hfill 
    \begin{subfigure}[t]{0.23\textwidth}
        \centering
        \includegraphics[width=\linewidth]{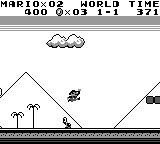}
    \end{subfigure}

    \vspace{1em}
    
    \begin{minipage}[t]{0.2\textwidth}\vspace{0pt}
        \includegraphics[width=\linewidth]{figures/example/example1/example1_base_step2.jpeg}
    \end{minipage}%
    \hfill
    \begin{minipage}[t]{0.78\textwidth}\vspace{0pt}
        \tiny
        <perception> Mario is standing on a platform at the bottom of the screen in World 1-1. There are two palm trees and a gray block to the left. To the right of Mario, there is a small enemy (likely a Koopa Troopa or similar). There are no coins or power-ups visible. The ground slopes upward to the right, and there is a cloud in the sky. No pipes or ladders are visible. </perception>
        
        <reasoning> 1. Mario is at the start of the level and needs to move forward to progress. 2. The enemy is to the right of Mario and appears to be stationary or moving slowly. Jumping over it or running past it is a safe option. 3. The ground slopes upward, which means running right will take Mario onto a higher platform, potentially avoiding the enemy and progressing. \textcolor{red}{4. There are no immediate obstacles in front of Mario that require jumping over, and no power-ups or coins to collect right now. 5. Pressing 'right' will move Mario forward while 'b' (hold for running) will allow him to move faster and safely pass the enemy. 6. Since no blocking obstacles are present, it is safe to run without needing to jump.} 7. While waiting ('noop') might be an option, it would slow progress. Therefore, moving forward is the best choice. </reasoning>
        
        <answer>['right', 'b']</answer>
    \end{minipage}

\vspace{1em}

    \begin{minipage}[t]{0.2\textwidth}\vspace{0pt}
        \includegraphics[width=\linewidth]{figures/example/example1/example1_rlsft_step1.jpeg}
    \end{minipage}%
    \hfill
    \begin{minipage}[t]{0.78\textwidth}\vspace{0pt}
        \tiny
        <perception>Mario is on the ground in World 1-1, with 2 lives and 3 coins. There is a coin floating above a small hill to the left, and a Goomba enemy is standing on the ground to the right. The screen shows a clear path ahead with no immediate obstacles.</perception>
        
        <reasoning> 1. Mario is currently on the ground and needs to progress forward. The immediate goal is to collect the coin and defeat the Goomba to continue. 2. The coin is floating above a small hill, which means Mario needs to jump to reach it. Pressing 'a' will allow Mario to jump and collect the coin. 3. After collecting the coin, Mario should move right to approach the Goomba. Pressing 'right' will move Mario toward the Goomba. \textcolor{blue}{4. To defeat the Goomba, Mario needs to jump on it or use a fireball if available. Since Mario is on the ground and the Goomba is standing, jumping on it is the most straightforward approach.} 5. Therefore, the best action is to jump to collect the coin and then move right to approach the Goomba.</reasoning>
        
        <answer>['a', 'right']</answer>
    \end{minipage}
    \caption{Example trajectories of base model (top) and \recipe{} (bottom). Base model fails in front of the enemy while \recipe{} successfully jumps over it. CoT shows that the base model continues to output \texttt{right} action, since it fails to accurately recognize the distance between Mario and enemy; however \recipe{} is able to jump at the right timing. \textcolor{red}{red} and \textcolor{blue}{blue} highlight the key differences in their generated CoTs.}
    \label{fig:example1}
\end{figure}

\begin{figure}
    \centering
    \begin{subfigure}[t]{0.23\textwidth}
        \centering
        \includegraphics[width=\linewidth]{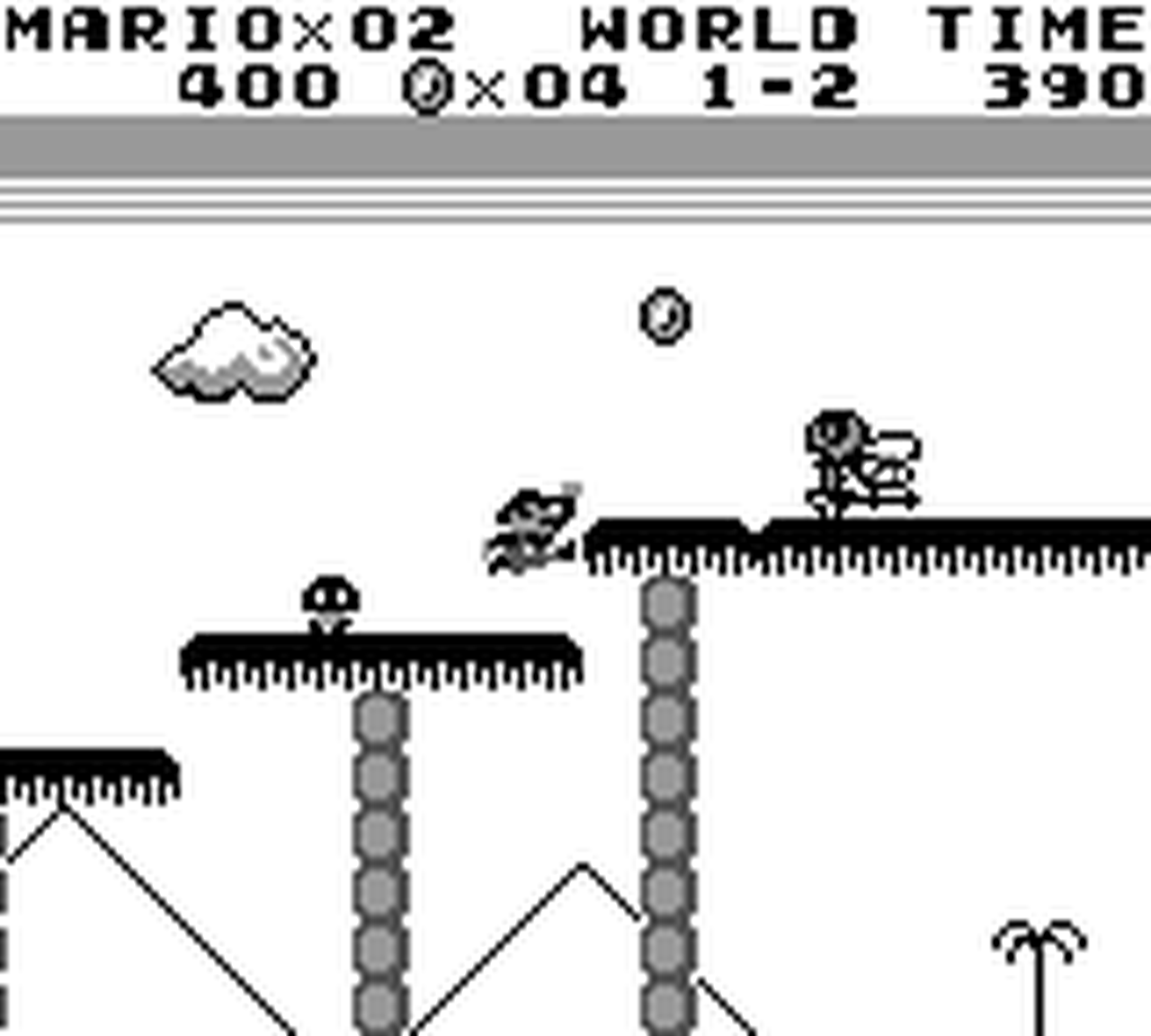}
    \end{subfigure}%
    \hfill
    \begin{subfigure}[t]{0.23\textwidth}
        \centering
        \includegraphics[width=\linewidth]{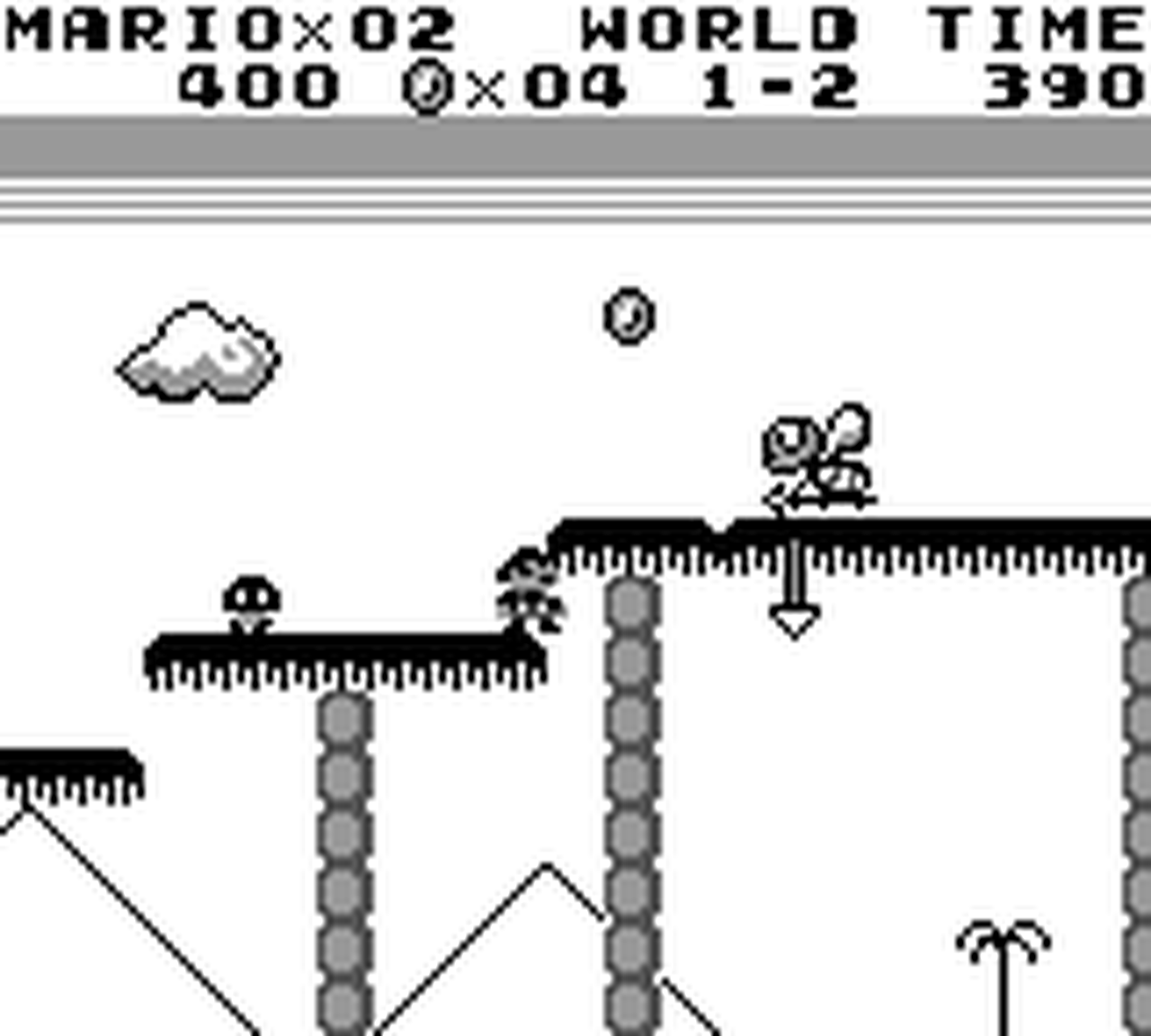}
    \end{subfigure}%
    \hfill 
    \begin{subfigure}[t]{0.23\textwidth}
        \centering
        \includegraphics[width=\linewidth]{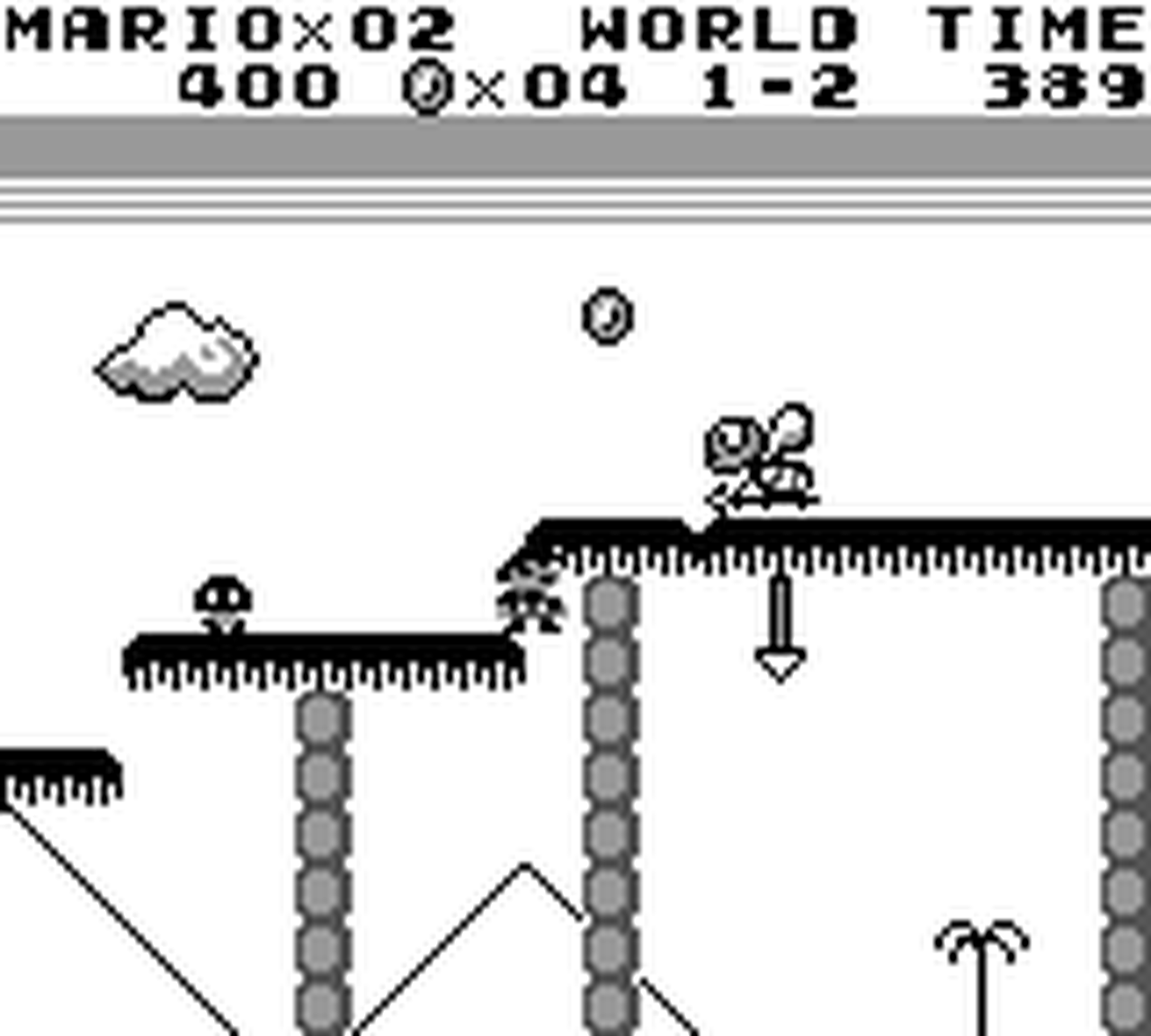}
    \end{subfigure}%
    \hfill 
    \begin{subfigure}[t]{0.23\textwidth}
        \centering
        \includegraphics[width=\linewidth]{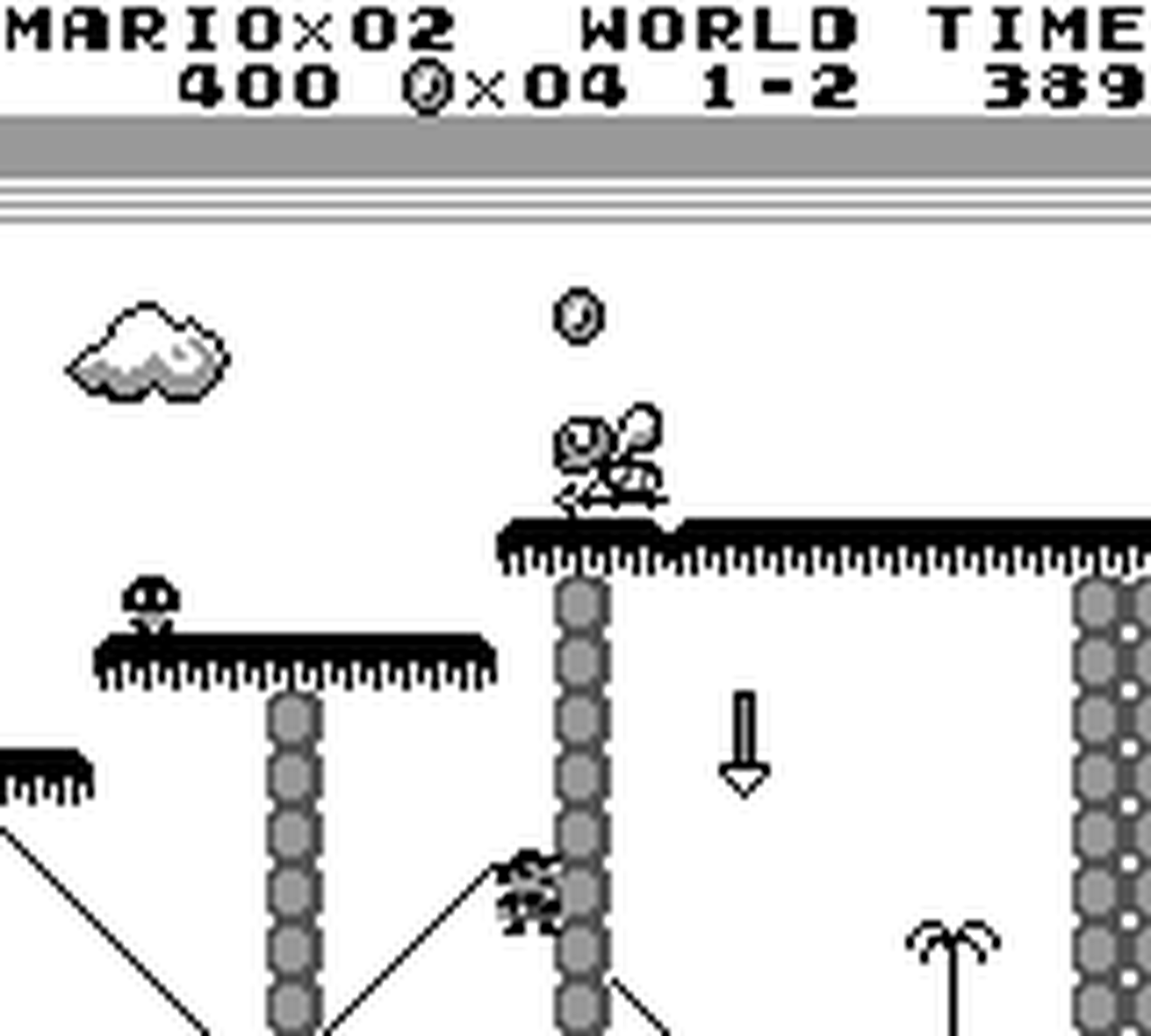}
    \end{subfigure}

    \vspace{1em}

    \begin{subfigure}[t]{0.23\textwidth}
        \centering
        \includegraphics[width=\linewidth]{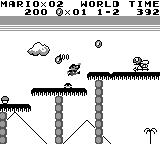}
    \end{subfigure}%
    \hfill 
    \begin{subfigure}[t]{0.23\textwidth}
        \centering
        \includegraphics[width=\linewidth]{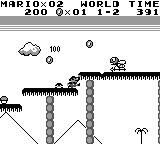}
    \end{subfigure}%
    \hfill 
    \begin{subfigure}[t]{0.23\textwidth}
        \centering
        \includegraphics[width=\linewidth]{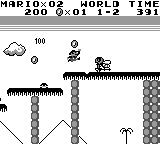}
    \end{subfigure}%
    \hfill 
    \begin{subfigure}[t]{0.23\textwidth}
        \centering
        \includegraphics[width=\linewidth]{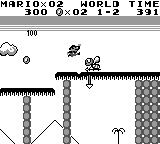}
    \end{subfigure}

    \vspace{1em}
    
    \begin{minipage}[t]{0.2\textwidth}\vspace{0pt}
        \includegraphics[width=\linewidth]{figures/example/example2/example2_base_step2.png}
    \end{minipage}%
    \hfill
    \begin{minipage}[t]{0.78\textwidth}\vspace{0pt}
        \tiny
        <perception>
Mario is standing on a ledge in World 1-2, with two Goombas on the platforms below and to the left. There is a coin floating above Mario. To the right, there is a pipe with a downward arrow indicating it can be entered. The screen also shows Mario has 2 lives and 4 coins, with 390 time remaining.
</perception>

<reasoning>
Mario is currently safe on the ledge, with no enemies directly below him. The coin above is within reach and can be collected by jumping. However, the immediate objective is to progress through the level. \textcolor{red}{The pipe on the right is the most direct way to advance, but it requires Mario to crouch or press down to enter it. Since the pipe is located to the right, Mario should move right to reach it. Pressing 'down' will allow him to enter the pipe if he steps on it, as this is standard behavior in Super Mario Land for pipe entry.} Therefore, the sequence should be to move right toward the pipe, and then press down to enter it, which will lead to the next part of the level. The coin can be collected later if needed, as progression is the priority.
</reasoning>

<answer>['right', 'down']</answer>
    \end{minipage}

\vspace{1em}
    \begin{minipage}[t]{0.2\textwidth}\vspace{0pt}
        \includegraphics[width=\linewidth]{figures/example/example2/example2_rlsft_step2.jpeg}
    \end{minipage}%
    \hfill
    \begin{minipage}[t]{0.78\textwidth}\vspace{0pt}
        \tiny
        <perception>
Mario is on a platform in World 1-2, with 2 lives and 1 coin. There are coins floating in the air and a Goomba enemy on the platform to the right. Mario is positioned on a lower platform, facing right, with a gap ahead that requires jumping to cross. There are vertical pillars between the platforms.
</perception>

<reasoning>
\textcolor{blue}{1. Mario needs to cross the gap to progress. The gap is wide, so Mario must jump to reach the next platform.}
2. There is a Goomba enemy on the platform to the right. If Mario jumps without avoiding it, he will take damage.
\textcolor{blue}{3. The best strategy is to jump while moving right to land on the next platform, avoiding the Goomba.}
\textcolor{blue}{4. Since Mario is on a platform, he can jump and move right simultaneously to cross the gap and land on the next platform.}
5. This will allow Mario to progress and collect the coins in the air once he lands.
</reasoning>

<answer>['a', 'right']</answer>
    \end{minipage}
    \caption{Example trajectories of base model (top) and \recipe{} (bottom). Base model falls into the gap while \recipe{} successfully jumps to cross the gap onto the higher platform. CoT shows that base model output \texttt{right} and \texttt{down} action, since it treats background scene object as the pipe available for entry; however \recipe{} is able to realize jumping onto the next platform is necessary to proceed in the game. \textcolor{red}{red} and \textcolor{blue}{blue} highlight the key differences in their generated CoTs.}
    \label{fig:example2}
\end{figure}

In \cref{fig:example1} and \cref{fig:example2}, we provide additional visualizations illustrating representative game scenarios. In each case, the base model fails to make progress, while \recipe{} successfully continues. We further highlight key differences in the generated chains of thought (CoTs), which shed light on the improved decision-making behavior of \recipe{}.

\end{document}